\documentclass[10pt,twocolumn,letterpaper]{article}

\usepackage{cvpr}

\usepackage{booktabs}
\usepackage{multirow}
\usepackage[most]{tcolorbox}
\usepackage{tikz}
\usetikzlibrary{positioning,arrows.meta}

\definecolor{cvprblue}{rgb}{0.21,0.49,0.74}
\usepackage[pagebackref,breaklinks,colorlinks,allcolors=cvprblue]{hyperref}

\def\confName{CVPR}
\def\confYear{2027}

\title{When a Larger Guidance Scale Repairs Text-to-Video Failures and When It Does Not}

\author{
   Taewon Kang$^{1}$, Matthias Zwicker$^{1}$ \\
   $^{1}$University of Maryland at College Park, United States \\
   {\large \texttt{taewon@umd.edu, zwicker@umd.edu}} \\
}

\definecolor{rxgreen}{RGB}{0,143,76}
\definecolor{rxred}{RGB}{190,40,40}
\newcommand{\rx}[1]{\textcolor{rxgreen}{\underline{#1}}}
\newcommand{\pd}{p^{\circ}}
\newcommand{\dem}{\mathcal{D}}

\begin{document}
\maketitle

\begin{abstract}
When a text-to-video diffusion model ignores part of a prompt, the usual first fix is to raise the classifier-free guidance scale. This paper is a measurement study of when that fix works. On Mochi, the same increase repairs failures on rare compositions (\emph{a snowy beach}) beyond what generating again does, does not repair negations (\emph{a highway with no vehicles}), where it fixes fewer failures than generating again does, and changes nothing for events that must happen late in a clip (\emph{a sandcastle that a wave washes flat}). Which of these happens can be known in advance, from the prompt and one inexpensive score. The score, which we call \emph{demand}, measures how strongly the denoiser reacts to the part of the prompt that the failed video leaves out, and it is read from the first quarter of denoising, before any video exists. Our main finding is a condition: raising the scale is effective only when the prompt contains no negation and no event tied to a specific time, and when demand is sufficiently high, with recovery clearly higher above a fitted threshold than below it. Published corrections for the three failure types, used as fixed probes, behave as this account predicts. A short decision rule follows: before generating again, a user can tell whether the most common fix is worth trying, and what to use when it is not.
\end{abstract}

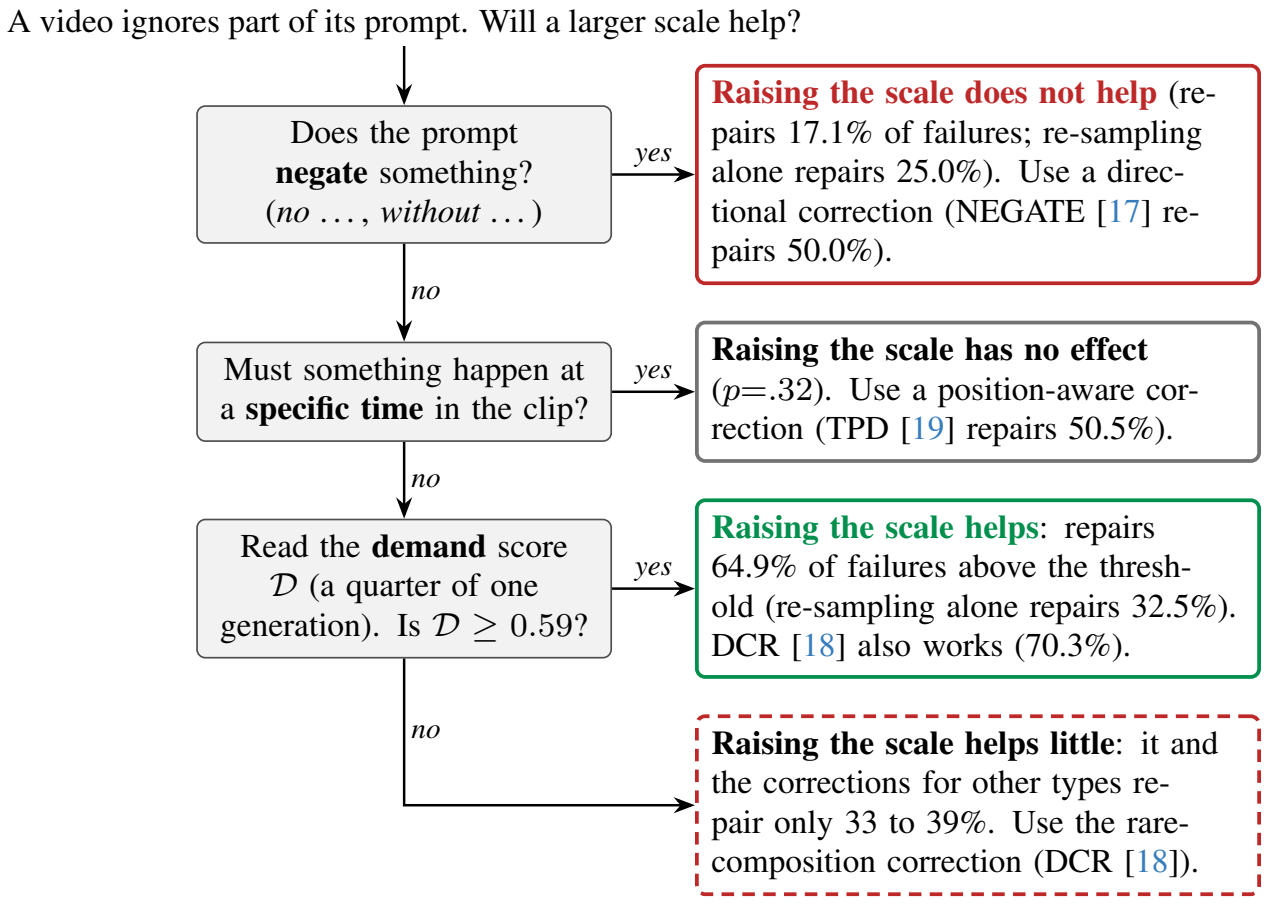
\begin{figure}[t]
\centering
\resizebox{\linewidth}{!}{%
\begin{tikzpicture}[font=\footnotesize, node distance=3mm and 7mm,
  q/.style={draw=black!70, rounded corners=2pt, align=center, text width=3.25cm, inner sep=3.5pt, fill=black!5},
  a/.style={draw=rxgreen, line width=0.9pt, rounded corners=2pt, align=left, text width=4.5cm, inner sep=3.5pt},
  harmbox/.style={draw=rxred, line width=0.9pt, rounded corners=2pt, align=left, text width=4.5cm, inner sep=3.5pt},
  flatbox/.style={draw=black!55, line width=0.9pt, rounded corners=2pt, align=left, text width=4.5cm, inner sep=3.5pt},
  n/.style={draw=rxred, line width=0.9pt, densely dashed, rounded corners=2pt, align=left, text width=4.5cm, inner sep=3.5pt},
  arr/.style={-{Stealth[length=1.8mm]}, semithick},
  lab/.style={font=\scriptsize\itshape, inner sep=1.5pt}]
\node[harmbox] (a1) {\textcolor{rxred}{\textbf{Raising the scale does not help}} (repairs 17.1\% of failures; re-sampling alone repairs 25.0\%). Use a directional correction (NEGATE~\cite{kang2026negate} repairs 50.0\%).};
\node[flatbox, below=of a1] (a2) {\textbf{Raising the scale has no effect} ($p{=}.32$). Use a position-aware correction (TPD~\cite{kang2026tpd} repairs 50.5\%).};
\node[a, below=of a2] (a3) {\textcolor{rxgreen}{\textbf{Raising the scale helps}}: repairs 64.9\% of failures above the threshold (re-sampling alone repairs 32.5\%). DCR~\cite{kang2026dcr} also works (70.3\%).};
\node[n, below=of a3] (a4) {\textbf{Raising the scale helps little}: it and the corrections for other types repair only 33 to 39\%. Use the rare-composition correction (DCR~\cite{kang2026dcr}).};
\node[q, left=of a1] (q1) {Does the prompt \textbf{negate} something?\\(\emph{no \dots}, \emph{without \dots})};
\node[q, left=of a2] (q2) {Must something happen at a \textbf{specific time} in the clip?};
\node[q, left=of a3] (q3) {Read the \textbf{demand} score $\dem$ (a quarter of one generation). Is $\dem \ge 0.59$?};
\draw[arr] ([yshift=5mm]q1.north) -- (q1.north);
\node[lab, anchor=south] at ([yshift=5mm]q1.north) {\normalfont\footnotesize A video ignores part of its prompt. Will a larger scale help?};
\draw[arr] (q1) -- node[lab, above] {yes} (a1);
\draw[arr] (q2) -- node[lab, above] {yes} (a2);
\draw[arr] (q3) -- node[lab, above] {yes} (a3);
\draw[arr] (q1) -- node[lab, right] {no} (q2);
\draw[arr] (q2) -- node[lab, right] {no} (q3);
\draw[arr] (q3.south) |- node[lab, pos=0.25, right] {no} (a4.west);
\end{tikzpicture}}
\caption{\textbf{When raising the guidance scale helps, and what follows in practice} (measured on Mochi; Table~\ref{tab:grid}, Fig.~\ref{fig:boundary}). The effect of the scale is this paper's finding; the alternatives in the first two boxes are published corrections on the failure type each was designed for.}
\label{fig:rule}
\end{figure}

\begin{table*}[t]
\centering
\small
\setlength{\tabcolsep}{5pt}
\begin{tabular}{@{}llll@{}}
\toprule
 & \textbf{Rare composition} & \textbf{Negation} & \textbf{Temporal} \\
\midrule
Request $p$ & \emph{a snowy beach with waves} & \emph{a highway at sunset with no vehicles} & \emph{a sandcastle on a beach where} \\
 & & & \emph{a wave washes it flat} \\
Default $\pd$ (made instead) & \emph{a tropical beach with waves} & \emph{a highway at sunset} & \emph{a sandcastle on a beach} \\
\midrule
Failure rate, untreated & .472 & .365 & .495 \\
Failure rate, scale raised & .297 & .407 & .527 \\
Effect of raising the scale & \textcolor{rxgreen}{\textbf{helps}} ($p<10^{-9}$) & \textcolor{rxred}{\textbf{does not help}} (.171 $<$ re-seed .250) & \textbf{no effect} ($p=.32$) \\
\midrule
Demand of failing prompts & lower than successes & higher than successes & lower than successes \\
Our reading & hard part barely registers & strong reaction, aimed at & hard part barely registers \\
 & & the forbidden content & and must be placed in time \\
\midrule
What repairs it (recovery) & any correction (.476 to .698), & NEGATE only (.500) & TPD only (.505) \\
 & incl.\ the raised scale (.540) & & \\
\bottomrule
\end{tabular}
\caption{\textbf{The paper at a glance.} Three failure types, the simpler prompt the model answers instead, the effect of the most common fix, what the demand score shows, and what works. All rates are on Mochi; $p$-values are exact McNemar tests against the untreated model. The re-seed rate is what generating again repairs without any method (Appendix~\ref{app:reseed}). Details are in Tables~\ref{tab:demand_main} and~\ref{tab:grid}.}
\label{tab:glance}
\end{table*}

\section{Introduction}
\label{sec:intro}

\begin{figure}[t]
    \centering
    \includegraphics[width=\linewidth]{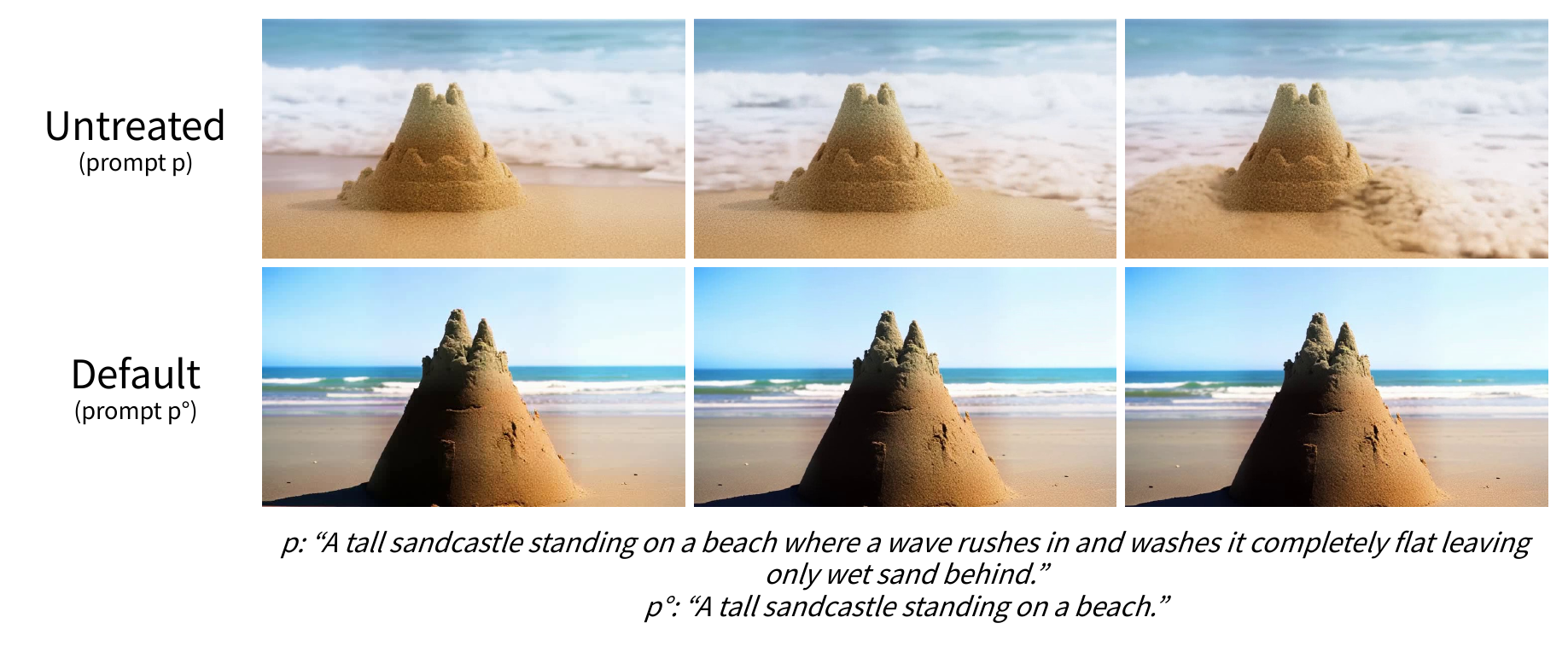}
    \caption{\textbf{A failed video is the video of the default prompt.} First, middle, and last frames. Top: the untreated clip for the full request $p$; the wave never arrives and the castle stands to the last frame. Bottom: a clip generated from the default prompt $\pd$ alone. Both rows show the same scene.}
    \label{fig:default}
\end{figure}

A user gives a text-to-video model the prompt \emph{``A tall sandcastle standing on a beach where a wave rushes in and washes it completely flat leaving only wet sand behind.''} The model returns a well-rendered beach, a well-rendered castle, and no wave (Fig.~\ref{fig:default}, top). We use this prompt as a running example and call it the sandcastle prompt. The usual next step is to raise the classifier-free guidance scale. In many cases the video that comes back is no better: the wave is still missing. This paper asks in which cases a larger scale does repair the video and in which it does not.

Whether raising the scale helps depends on the kind of failure (Table~\ref{tab:glance}). On Mochi, it lowers the failure rate on rare compositions such as \emph{a snowy beach} from 47.2\% to 29.7\%. On negations such as \emph{a highway with no vehicles} it repairs 17.1\% of the failures, fewer than the 25.0\% that generating again with a new seed repairs. For the sandcastle, whose wave must arrive late in the clip, it changes nothing (49.5\% to 52.7\%, $p=.32$). Many inference-time methods modify the guidance update to repair such failures~\cite{kang2026negate, kang2026dcr, kang2026tpd, brack2023sega, chefer2023attend, liu2022compositional, zhang2025constrained}. Each of these methods was built to repair one particular kind of failure, for example only negations, and its authors tested it only on prompts of that kind. The plain scale, which is what a user reaches for first, has not been examined in the same way. Our answer has two parts. First, raising the scale repairs some failures and not others. Second, a user does not have to try it to find out which will happen: reading the prompt, and computing one number from the model at a fraction of the cost of one generation, is enough to tell beforehand.

Why does the same change help in one case and not in the other two? The three cases share one cause. The model does not make a random bad video; it makes a good video of an easier prompt than the one it was given. For the sandcastle, the simpler prompt is \emph{a sandcastle on a beach}, the request with its wave deleted, and that is the video the user received. The other two examples come back as videos of \emph{a tropical beach} and of \emph{a highway at sunset}. We call the request with its difficult part deleted the \emph{default prompt} $\pd$, and we verify that a failed video for $p$ looks like the video of $\pd$ by generating from $\pd$ directly (Fig.~\ref{fig:default}).

The default prompt gives us something to measure against. The full request and the default prompt differ only in the missing part, so if we give the model both and compare what it predicts for each, the difference is the model's reaction to that part. For the sandcastle, we ask the denoiser for its prediction under the full request and under \emph{a sandcastle on a beach}, on the same noisy latent, and take the size of the difference relative to the size of the whole text-conditioned update. We call this number \emph{demand}. It says how strongly the model reacts to the wave, and it needs no judge and no finished video, since the first quarter of the denoising steps suffices.

Demand separates two ways of failing, and they account for the three outcomes. Negation prompts that fail have \emph{higher} demand than negation prompts that succeed: the model reacts strongly to \emph{no vehicles}, but toward the vehicles, a reaction that raising the scale cannot turn around. For rare-composition and temporal prompts it is the other way round: those that fail have \emph{lower} demand than those that succeed. Low demand means that the model's prediction hardly changes when the difficult part is added, that is, the model almost ignores it. A larger scale multiplies the influence of the text, so it can strengthen this weak reaction until the difficult part appears in the video, which is why the scale helps rare compositions. It still does not help the sandcastle, whose demand is also low, and the reason can be read off the prompt: the wave must appear late in the clip, and the scale is a single number applied equally to every frame, so it cannot make something happen at one time and not at another.

Published corrections let us test this explanation. One exists for each failure type: NEGATE~\cite{kang2026negate} steers away from forbidden content, DCR~\cite{kang2026dcr} steers away from the common counterpart of a rare combination, and TPD~\cite{kang2026tpd} acts along the frame axis. We apply all three, together with the raised scale, to all three failure types, each in its published configuration. We do not use these runs to rank the corrections; each is compared with the unmodified model (Table~\ref{tab:grid}). That each correction repairs the failure type it was designed for is expected and is not a result of this paper. The new information is in the other cells, where a method meets a failure type it was not designed for, and in the rows for the raised scale: on Mochi the raised scale repairs few negation failures where the directional correction repairs half, and demand relates to failure in opposite ways in negation and in the other two failure types, so no single scalar fix can serve all three.

Our contributions are the following. (1)~\emph{The main finding.} On Mochi, raising the guidance scale repairs failures only under three conditions: the prompt contains no negation; it does not require something to happen at a particular time in the clip; and the model's demand score for the prompt is high enough. Above a threshold that we fit to the data, the scale repairs about 1.7 times as large a share of failures as below it. (2)~\emph{An explanation.} A failed video is the video of the default prompt, which we confirm by generating from the default prompt directly. (3)~\emph{A measurement.} The demand score, read from the model before any video is generated. It predicts which rare-composition prompts will fail, whereas three predictors computed from the prompt text alone (its length, a rating of how rare the scene is, and the distance between the two prompts in the text encoder) do no better than random guessing on two of the three models. (4)~\emph{A practical implication.} The decision rule of Fig.~\ref{fig:rule}. Two of its branches send negations and temporal prompts to the corrections that were designed for them, which is not new; what is new is the branch for the plain scale, which says when to raise it and when to leave it alone. The conditions are measured on Mochi, and on a second model they come out differently (Appendix~\ref{app:cogscale}), so they are re-measured, not assumed, on the model in use.

\section{Related Work}
\label{sec:related}

Classifier-free guidance~\cite{ho2022classifier} has spawned many training-free controls that reshape its update, but none of this work says when the plain scale itself helps. The controls include semantic directions~\cite{brack2023sega, kang2023semantic}, attention-based methods~\cite{chefer2023attend, feng2022training, rassin2023linguistic}, composable forms~\cite{liu2022compositional}, concept erasure~\cite{gandikota2023erasing, gandikota2024unified}, constrained sampling~\cite{zampini2025training, zhang2025constrained}, and corrections to the update itself~\cite{chung2024cfg++, si2024freeu, bansal2023universal, xie2025dymo}. Pathologies of the scale are documented, including spatial inconsistency of a single scale~\cite{shen2024rethinking} and amplification of unintended attributes under any global increase~\cite{xia2025factored}, which we observe as color saturation. To test our explanation we use published corrections as fixed probes, selected by three criteria: they act only in the guidance update, they report a validated configuration on Mochi~\cite{mochi}, and they ship an evaluation suite for one failure type. NEGATE, DCR, and TPD~\cite{kang2026negate, kang2026dcr, kang2026tpd} satisfy all three; we know of no others that do.

Each of the three failure types has its own literature, on compositionality, on negation, and on temporal control, and several recent works use signals from inside the model as measurements, as we do. We review both in the supplementary material (Appendices~\ref{app:related} and~\ref{app:scope}).

\section{Setup}
\label{sec:experiments}

\paragraph{Failure types, models, and judging.}\label{subsec:protocol}
We study three failure types, each of which comes with a published test suite and a published correction. In \emph{rare composition} the prompt combines things that seldom co-occur, and the model renders the common version. In \emph{negation} the prompt excludes something, and the model renders it anyway. In \emph{temporal} prompts, of which the sandcastle is one, an event must emerge late in the clip, and it never arrives. Each suite consists of short English prompts in eight categories that vary where the difficulty sits (prompt counts and category definitions in Appendix~\ref{app:suites}). We use the suites unmodified. Every prompt was written to provoke failure, so the rates below overstate failure rates in ordinary use; in exchange, the suites supply the variation that an explanation has to account for. We generate one clip per prompt on Mochi~\cite{mochi}, HunyuanVideo~\cite{kong2024hunyuanvideo}, and CogVideoX~\cite{yang2025cogvideox}, each with its default sampling settings. Each suite's original judge labels every clip as a \emph{failure} or a success, and the failure rate is the fraction of failures.

\paragraph{Corrections and the raised scale.}\label{subsec:treatment}
We apply four methods to failing prompts, all on Mochi, where the three published corrections were validated by their authors, in their published settings and with nothing tuned. NEGATE~\cite{kang2026negate}, designed for negation, steers the update away from the forbidden content. DCR~\cite{kang2026dcr}, designed for rare composition, steers away from the common counterpart. TPD~\cite{kang2026tpd}, designed for temporal prompts, corrects along the frame axis, so it can place content late in the clip. The fourth method is the raised scale itself, labeled scale-up in the tables: plain classifier-free guidance with a larger scale, which makes the update stronger without changing its direction and therefore shows what strength alone can do. We apply all four methods to every failure type. Each correction takes a reference prompt that says what to steer away from, and outside the failure type it was designed for we fill that slot with the default prompt (Sec.~\ref{subsec:default_operator}, Appendix~\ref{app:grid}).

\paragraph{Measures.}
For each prompt we compare two clips: the \emph{untreated} clip, generated with standard settings, and the \emph{treated} clip, generated with a method applied. We report two fractions. \emph{Recovery} is computed over the prompts whose untreated clip failed, and is the fraction of them whose treated clip succeeds. \emph{Induction} is computed over the prompts whose untreated clip succeeded, and is the fraction of them whose treated clip fails; it measures the damage a method does to prompts that were fine without it. When we say that a method \emph{repairs} a share of failures, we mean its recovery. Significance is an exact McNemar test on the pairs that changed. A method is compared with the untreated model and not with another method, with one exception that the decision rule requires (Sec.~\ref{par:rule}). The full protocol is in the supplementary material (Appendices~\ref{app:formal} and~\ref{app:grid}).

\begin{figure}[t]
    \centering
    \includegraphics[width=1.0\linewidth]{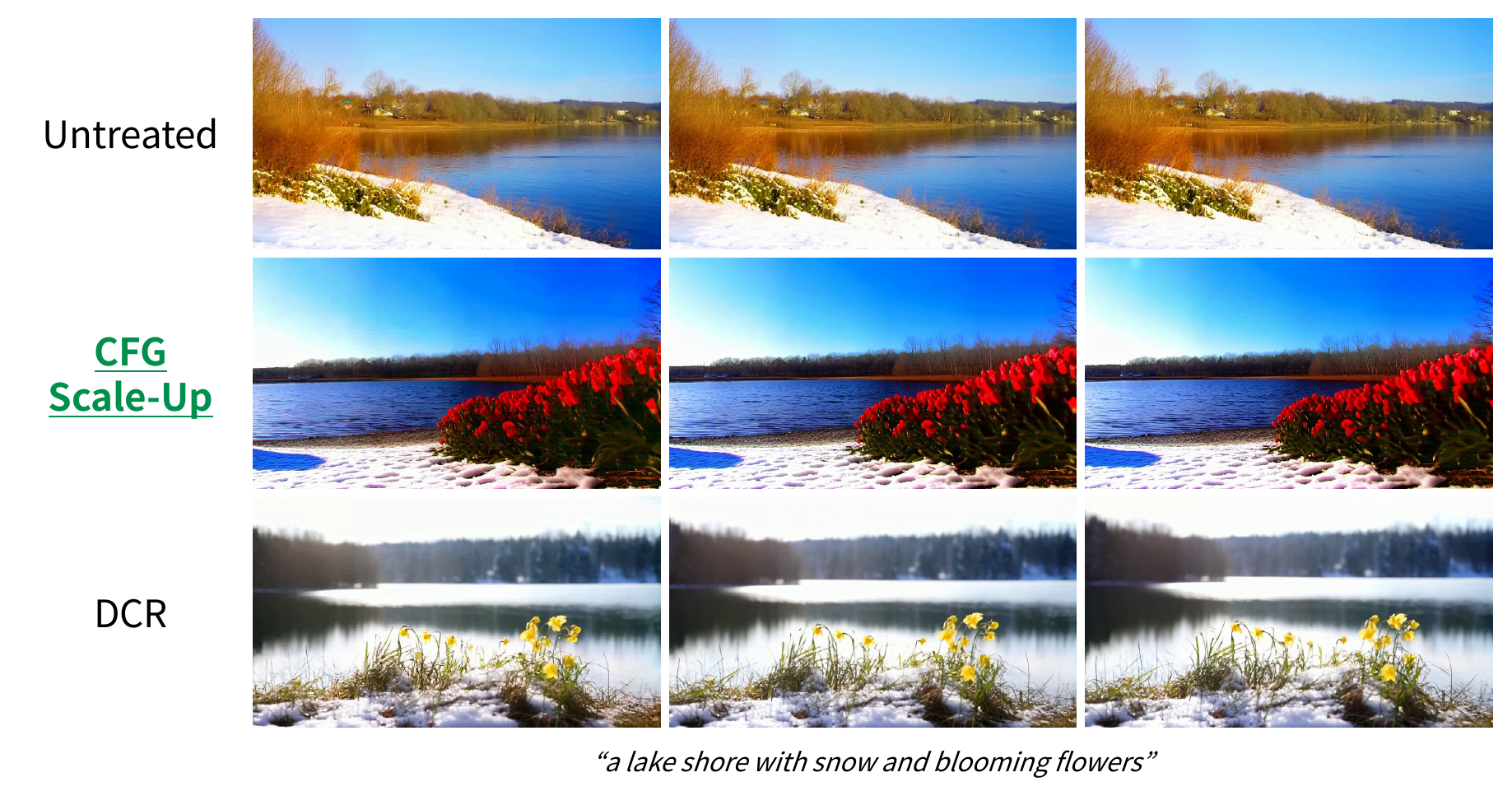}
    \caption{\textbf{Rare composition: a larger scale alone can be enough.} In Figs.~\ref{fig:rx_rarecomp} to~\ref{fig:rx_temporal}, a \rx{green underlined} label marks the method the rule selects. This prompt's demand (0.73) is above the threshold ($D^{*}=0.59$). Middle: a larger scale and nothing else brings snow and blossoms together (54\% of failures repaired). Bottom: DCR reaches the same outcome with a natural palette (recovery .698); the raised scale saturates colors, because it amplifies every attribute at once~\cite{xia2025factored}.}
    \label{fig:rx_rarecomp}
\end{figure}

\begin{figure}[t]
    \centering
    \includegraphics[width=1.0\linewidth]{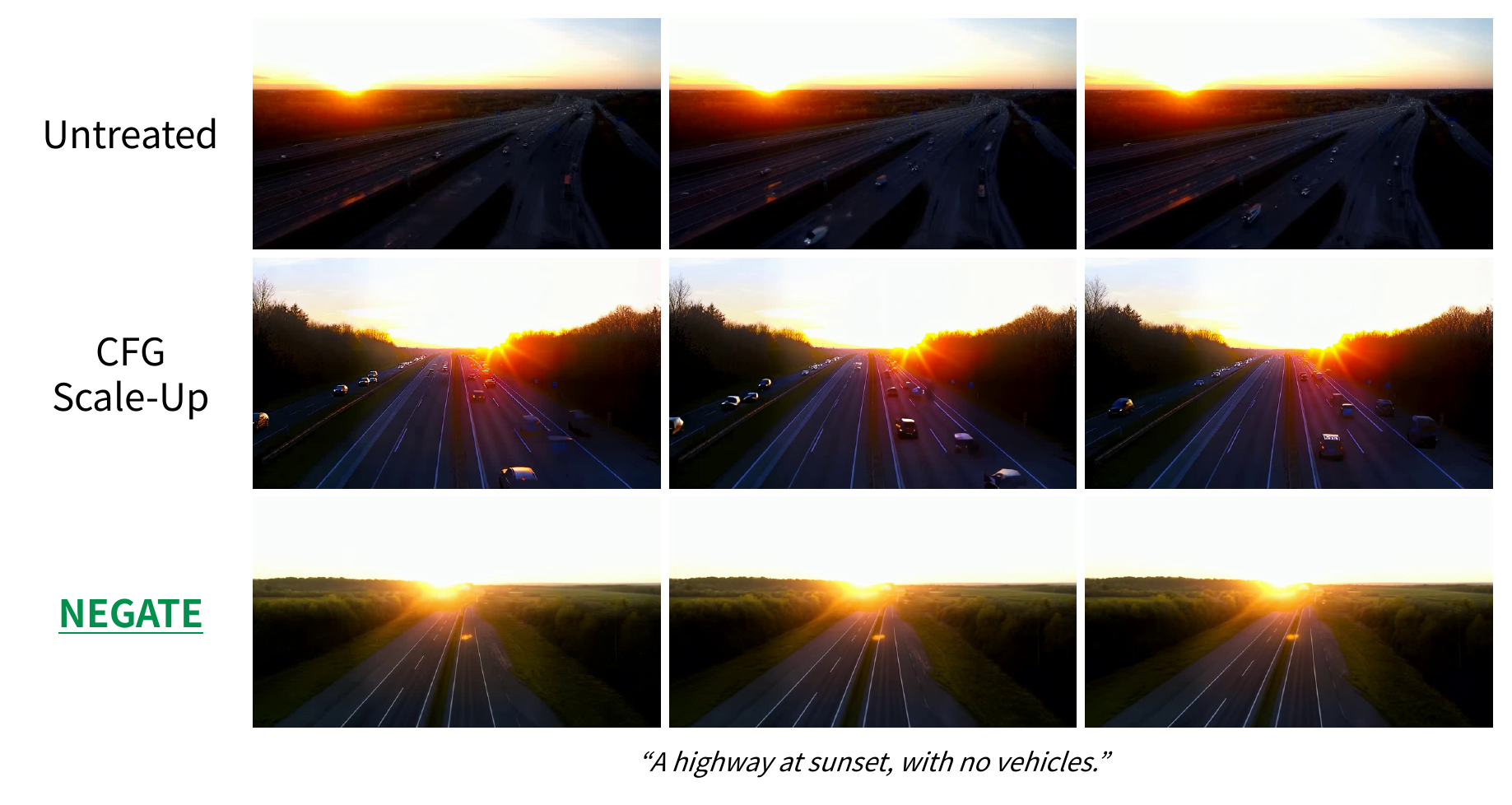}
    \caption{\textbf{Negation: raising the scale does not remove the vehicles; NEGATE does.} Middle: with a larger scale the vehicles persist. Bottom: NEGATE returns the same sunset road without vehicles (50\% of failures repaired).}
    \label{fig:rx_negation}
\end{figure}

\section{Finding: One Fix, Three Outcomes}
\label{sec:finding}

\begin{figure}[t]
    \centering
    \includegraphics[width=1.0\linewidth]{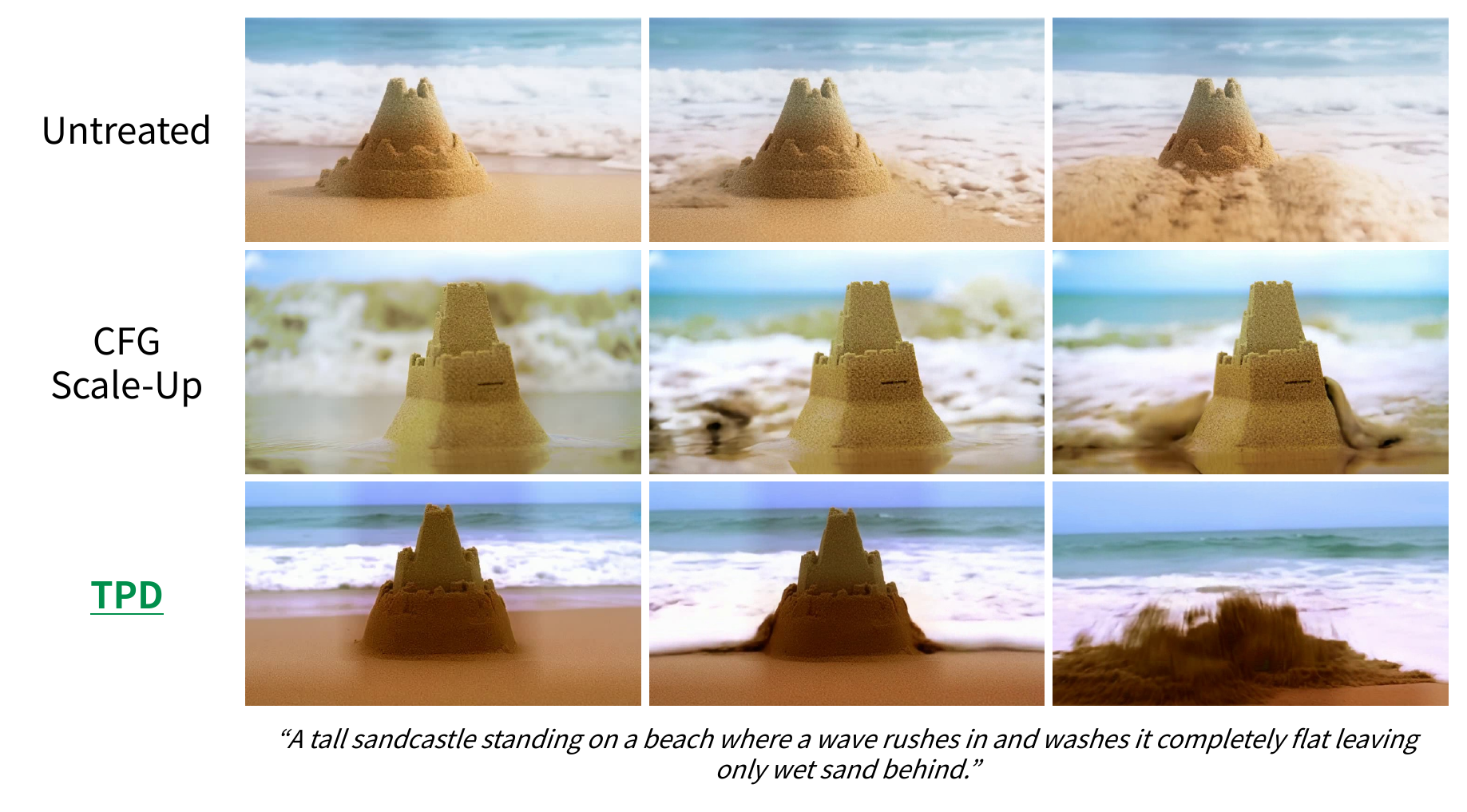}
    \caption{\textbf{Temporal: only the position-aware correction delivers the event.} Top: untreated, the castle stands to the last frame. Middle: a larger scale re-renders the same early scene. Bottom: TPD acts along the frame axis, and the wave arrives and flattens the castle (51\% of failures repaired).}
    \label{fig:rx_temporal}
\end{figure}

Raising the guidance scale is one operation, but on Mochi its outcome differs between the three failure types (Table~\ref{tab:glance}; scale-up rows of Table~\ref{tab:grid}). In rare composition it helps: the failure rate falls from .472 to .297, and 54.0\% of untreated failures are repaired ($p<10^{-9}$). In Fig.~\ref{fig:rx_rarecomp}, snow and blossoms appear together once the scale is raised. In negation it does not help: it repairs only 17.1\% of the failures, fewer than the 25.0\% that simply generating again repairs, the failure rate does not fall (.365 to .407), and the vehicles that should be absent in Fig.~\ref{fig:rx_negation} remain. For temporal prompts it does nothing: the rate moves from .495 to .527 ($p=.32$), because the clips it repairs are offset by clips it breaks, and in Fig.~\ref{fig:rx_temporal} the sandcastle is re-rendered and still stands.

Raising the scale is therefore not a safe default. Whether it helps depends on the failure type, and the same three outcomes hold across repeated seeds and across a range of scales (Appendices~\ref{app:seeds} to~\ref{app:dose}). Any explanation has to produce all three at once, and the next section gives one.

\section{Explanation: The Model Answers a Simpler Prompt}
\label{sec:method}

\subsection{The default prompt}
\label{subsec:default_operator}

A failed video is the video of the request with its difficult part deleted. The sandcastle request contains one specification that the model's training distribution disprefers, the wave that flattens the castle late in the clip, and the model returns the video of what remains, \emph{a sandcastle on a beach}. We call what remains the default prompt $\pd$. Every prompt in the three suites has one such specification, and $\pd$ is built by deleting it and keeping the rest of the scene. For a temporal prompt we cut the late event, as above. For a rare composition we relax the rare factor, so that \emph{a snowy beach with waves} becomes \emph{a tropical beach with waves}. For a negation we delete the negation and its scope, so that \emph{a highway at sunset with no vehicles} becomes \emph{a highway at sunset}. In each case $\pd$ is a coherent prompt and a more probable one than $p$, and because the construction edits text only, one $\pd$ serves every model.

Generating from the default prompt alone reproduces the failed video. Figure~\ref{fig:default} shows this for the sandcastle: the clip for the full request and the clip for \emph{a sandcastle on a beach} show the same scene. With one $\pd$ clip per category, the judge finds the default realized in 8/8 temporal cases and 6/8 rare-composition cases. For negation the check is uninformative (4/8), because a $\pd$ clip can satisfy the negation by accident (an empty highway at sunset). As a second check we use the CLIP image-text model to compare how similar each $\pd$ clip is to the text of $\pd$ and to the text of its source prompt $p$. In rare composition the $\pd$ clips are more similar to $\pd$ than to $p$ (by $+.028$). In negation the two similarities do not differ, as expected, because CLIP is insensitive to negation words~\cite{alhamoud2025vision}. 

\subsection{Demand: how strongly the model reacts to the hard part}
\label{subsec:demand}

Demand measures how much the denoiser's prediction changes when the difficult part is added back to the default prompt. Let $\epsilon_\theta(x_t, t, c)$ denote the noise that the frozen denoiser of model $\theta$ predicts for the noisy latent $x_t$ at denoising step $t$ under text conditioning $c$, and let $\Phi_\theta$ be the model's text encoder. For a prompt $p$ and its default $\pd$ we use three predictions, all on the \emph{same} latent $x_t$ of an ordinary, untreated sampling run:
\begin{equation*}
\begin{aligned}
\epsilon_{\text{text}} &= \epsilon_\theta\big(x_t, t, \Phi_\theta(p)\big) && \text{(full prompt)},\\
\epsilon^{\circ} &= \epsilon_\theta\big(x_t, t, \Phi_\theta(\pd)\big) && \text{(default prompt)},\\
\epsilon_{\text{uncond}} &= \epsilon_\theta\big(x_t, t, \varnothing\big) && \text{(no text)}.
\end{aligned}
\end{equation*}
Classifier-free guidance~\cite{ho2022classifier} moves the sample along $\epsilon_{\text{text}} - \epsilon_{\text{uncond}}$, multiplied by the guidance scale $w$. Demand is
\begin{equation}
\dem(p;\theta) \;=\; \frac{1}{|\mathcal{T}|}\sum_{t \in \mathcal{T}} \frac{\|\epsilon_{\text{text}} - \epsilon^{\circ}\|_2}{\|\epsilon_{\text{text}} - \epsilon_{\text{uncond}}\|_2},
\label{eq:demand}
\end{equation}
averaged over a set $\mathcal{T}$ of denoising steps. Demand is the only formula the main text needs; the formal definitions of every other quantity are given in the supplementary material (Appendix~\ref{app:formal}).

The ratio has a simple reading. Its numerator is the change in the prediction when the difficult part, the wave in our example, is added back to the default, and its denominator is the whole change the text makes, which removes model-specific magnitudes. Two extreme cases show how to read the number. If demand is close to 0, the model predicts almost the same thing whether or not the wave is in the prompt, so the wave has almost no influence on the video. If demand is close to 1, mentioning the wave changes the prediction about as much as the entire prompt does, so the model reacts to the wave strongly. Demand is a size, like a distance, and is therefore never negative. It says how much the prediction changes and not whether the change goes toward what the prompt asks for or away from it.

\label{subsec:exp_localization}Demand is also cheap. It uses no judge and no reference video, and the early steps are enough: averaged over only the first quarter of the steps, it ranks prompts almost as the full trajectory does (Spearman $\rho = 0.876/0.871/0.857$ on Mochi/HunyuanVideo/CogVideoX). The price is a quarter of one generation's denoiser passes plus two text encodings. The early steps suffice because that is when the outcome is settled: on Mochi the demand gap between failing and succeeding rare-composition prompts is already at full depth within the first quarter of denoising and closes by the final quarter (Appendix~\ref{app:localization}). We use this first-quarter demand everywhere below.

\subsection{Two ways to fail}
\label{subsec:exp_demand}

\begin{table}[t]
\centering
\footnotesize
\setlength{\tabcolsep}{5pt}
\begin{tabular}{@{}lccc@{}}
\toprule
\multicolumn{4}{@{}l}{\emph{(a) Correlation between demand and failure, within each failure type}} \\
 & Negation & Rare comp. & Temporal \\
\midrule
Mochi        & $+0.092$ & $-0.176$ & $-0.116$ \\
HunyuanVideo & $+0.290$ & $-0.345$ & $-0.267$ \\
CogVideoX    & $+0.148$ & $-0.070$ & $-0.043$ \\
\midrule
\multicolumn{4}{@{}l}{\emph{(b) AUROC gained by adding demand to all text-only predictors}} \\
 & One slope & Slope per failure type & Rare comp. only \\
\midrule
Mochi        & $+0.025$ & $+0.050$ & $+0.040$ \\
HunyuanVideo & $+0.056$ & $+0.104$ & $+0.035$ \\
CogVideoX    & $+0.001$ & $+0.011$ & $+0.049$ \\
\bottomrule
\end{tabular}
\caption{\textbf{What demand shows.} (a)~Point-biserial correlation between failure and demand. Failing negations have higher demand than succeeding ones; failing rare-composition and temporal prompts have lower demand. The direction is the same on all three models. (b)~Five-fold cross-validated gain in AUROC (Sec.~\ref{sec:checks}); the last column uses category fixed effects. Full tables are in Appendix~\ref{app:tables}.}
\label{tab:demand_main}
\end{table}

For each failure type we compare the demand of the prompts that fail with the demand of the prompts that succeed (Table~\ref{tab:demand_main}a). The comparison comes out one way for negation and the opposite way for the other two failure types, on all three models.

\emph{In negation, the prompts that fail are those the model reacts to more.} Their demand is higher than that of the negation prompts that succeed (correlations $+0.092$ to $+0.290$). This seems odd at first: the model clearly notices \emph{no vehicles}, and yet the vehicles are in the video. Both facts hold if the reaction goes toward the vehicles instead of away from them, so that mentioning vehicles makes them more likely even after the word \emph{no}. A larger scale cannot turn that reaction around, consistent with the low recovery of the raised scale on negation (Sec.~\ref{sec:finding}).

\emph{In rare composition and in temporal prompts, the prompts that fail are those the model reacts to less.} Their demand is lower than that of the prompts that succeed ($-0.043$ to $-0.345$). The difficult part changes the prediction very little, so the video stays close to the default. A larger scale strengthens this weak reaction, and in rare composition that is often enough for the rare element to appear. A plausible cause, which we did not measure, is that the model saw few examples of the requested combination in training.

Temporal prompts are the exception: their demand is also low when they fail, and yet the scale does not help them. The reason is timing. The sandcastle's wave must arrive late in the clip, and the scale is one number applied equally to every frame, so it can make the text count for more but cannot make the wave arrive late. The model shows the same timing. Computed frame by frame, the demand of temporal prompts is higher in the later frames, where their events belong, on two of the three models; on HunyuanVideo the rise starts near the middle of the clip, close to the average event position $\bar{\gamma}=0.49$ (Appendix~\ref{app:localization}).

A prompt can therefore fail in two ways: the model reacts strongly, in what behaves as the wrong direction (negation), or it reacts weakly (rare composition and temporal prompts). Making the update stronger helps only in the second case, and only when nothing has to be placed in time. The correlations are small; the explanation rests on their sign, which is opposite for negation and the same on every model.

\begin{table}[t]
\centering
\scriptsize
\setlength{\tabcolsep}{4pt}
\begin{tabular}{lcccc}
\toprule
Method & Failure rate & Recovery & Induction & McNemar $p$ \\
\midrule
\multicolumn{5}{l}{\textit{Negation, untreated .365}} \\
NEGATE$^{\dagger}$ & .230 & .500 & .075 & $<10^{-6}$ \\
DCR & .365 & .240 & .138 & $1.0$ \\
TPD & .357 & .253 & .134 & $.81$ \\
Scale-up & .407 & .171 & .165 & $.05$ \\
\midrule
\multicolumn{5}{l}{\textit{Rare composition, untreated .472}} \\
DCR$^{\dagger}$ & .295 & .698 & .289 & $<10^{-6}$ \\
NEGATE & .328 & .476 & .152 & $<10^{-6}$ \\
TPD & .325 & .519 & .185 & $<10^{-6}$ \\
Scale-up & .297 & .540 & .152 & $<10^{-9}$ \\
\midrule
\multicolumn{5}{l}{\textit{Temporal, untreated .495}} \\
TPD$^{\dagger}$ & .323 & .505 & .153 & $<10^{-6}$ \\
NEGATE & .502 & .263 & .272 & $.85$ \\
DCR & .453 & .333 & .243 & $.14$ \\
Scale-up & .527 & .328 & .386 & $.32$ \\
\bottomrule
\end{tabular}
\caption{\textbf{Every method on every failure type} (Mochi). $^{\dagger}$: the correction designed for that failure type. Each cell is scored by the failure type's original judge and compared with the untreated model only: failure rate, paired recovery and induction, exact McNemar test (protocol in Appendix~\ref{app:grid}).}
\label{tab:grid}
\end{table}

\section{Testing the Explanation}
\label{sec:testing}

\subsection{What the explanation predicts for each method}
\label{subsec:exp_treatment_label}

If the explanation is right, it says in advance how each of the four methods behaves on each failure type. A correction repairs the failure type it was designed for by construction, so the informative cells of Table~\ref{tab:grid} are the others, and they bear out all three predictions. For negation, only a correction that steers away from the forbidden content should work. NEGATE does: of the negation prompts that failed untreated, 50.0\% succeed once NEGATE is applied, and it breaks few prompts that had succeeded (induction .075). The two corrections designed for other failure types leave the failure rate where it was (net change 0.000 at $p=1.0$ for DCR, $-.008$ at $p=.81$ for TPD), and the raised scale does not lower it either. For rare composition, where the signal is present but weak, any strengthening should work, and all four methods do lower the failure rate, by .145 to .178 (all $p<10^{-6}$). For temporal prompts, only a correction that knows about frame position should work: TPD repairs 50.5\% of failures, and no other method changes the rate significantly ($p=.85$, $.14$, $.32$).

The pattern can be stated plainly. Each failure type is missing something different, and a method helps only if it provides that thing. A negation failure behaves as a reaction in the wrong direction, so it needs a method that changes the direction of the update (NEGATE). A rare-composition failure is a reaction that is too weak, so it needs a stronger update, which all four methods provide. A temporal failure is an event that is not placed in time, so it needs a method that treats early and late frames differently (TPD). Raising the scale only makes the update stronger. It changes neither the direction nor the timing, which is why it helps rare compositions and nothing else (Figs.~\ref{fig:rx_rarecomp} to~\ref{fig:rx_temporal}).

\begin{figure*}[t]
    \centering
    \includegraphics[width=1.0\textwidth]{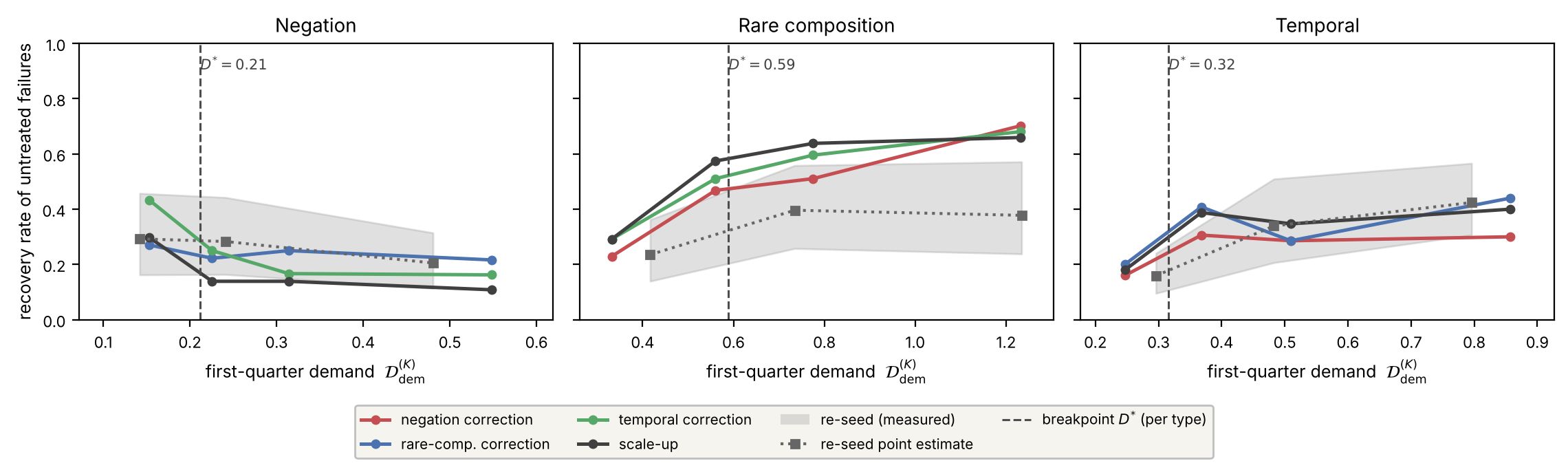}
    \caption{\textbf{Recovery against demand.} Fraction of untreated failures repaired, by quartile of the failures' demand, for the methods \emph{not} designed for each failure type (in the legend, the negation, rare-composition, and temporal corrections are NEGATE, DCR, and TPD). Dashed line: fitted threshold. Gray band: what generating again repairs without any method. Only in rare composition do the methods rise above the band as demand grows, from the threshold $D^{*}=0.59$ on (details in Appendix~\ref{app:reseed}).}
    \label{fig:boundary}
\end{figure*}

\subsection{How far the scale reaches}
\label{subsec:exp_boundary}

Demand was defined to explain why prompts fail, and it also tells which failures the scale can repair. Figure~\ref{fig:boundary} sorts the rare-composition failures by demand into four groups of equal size and plots how often each group is repaired. In the group with the lowest demand the raised scale repairs about a quarter of the failures, no more than generating again would. In the group with the highest demand it repairs about two thirds. The two corrections built for other failure types, NEGATE and TPD, show the same rise (.23 to .29 in the lowest group and .66 to .70 in the highest, over the three methods; $r = +0.22$ to $+0.31$, $p \le 0.002$), so the rise belongs to the prompts and the model and not to one method. Most of it happens at low demand, after which recovery levels off. Fitting two straight lines that meet at one point puts the bend at $D^{*} = 0.59$, with a wide confidence interval ($[0.49,\, 1.14]$; Table~\ref{tab:segmented}), and we use this value as an operating point rather than a sharp boundary. When the model hardly reacts to the difficult part, a stronger update has almost nothing to strengthen.

Negation and temporal prompts show no such gain. There the methods not built for the failure type repair about as much as generating again does, at every level of demand (gray bands in Fig.~\ref{fig:boundary}). Demand therefore enters the decision in one place, the rare compositions that the scale can reach. It also changes how a negative result should be read: when the scale fails on a low-demand prompt, that says little about the scale, because such prompts are out of reach for any method that only strengthens the update. The correction built for rare composition, DCR, is the exception and keeps working at low demand. The full analysis is in Appendix~\ref{app:reseed}.

\subsection{Practical implication: a decision rule}
\label{par:rule}

Taken together, the results say when raising the guidance scale is worth trying on Mochi. The failing prompt must contain no negation, it must place no event at a specific time in the clip, and its demand must clear $D^{*}=0.59$. Above the threshold the raised scale repairs .649 of the failures. Below it the scale repairs .385, close to what generating again repairs, and DCR is the better choice. Above the threshold both work, and the scale is preferable because it breaks fewer prompts that had succeeded (induction .132 against .276; Appendix~\ref{app:reseed}).

Figure~\ref{fig:rule} arranges these conditions as three questions that a user answers in order after a video has failed. Does the prompt say that something should be absent (\emph{no}, \emph{without})? Then the scale should not be raised, and a correction that steers away from the unwanted content applies (NEGATE repairs 50.0\% of failures). Does it require something to happen at a particular time in the clip? Then raising the scale will not help, and a correction that can act on the late frames applies (TPD, 50.5\%). Otherwise, does the demand score reach $D^{*}$? If so, raise the scale; if not, use the rare-composition correction (DCR), and only if that fails too generate again or switch models (Appendix~\ref{app:switch}). For the sandcastle, the second question applies: leave the scale alone and use the correction that places the wave late in the clip.

Only the third question needs the model, and the demand score costs about a quarter of the computation of one video. The rule is measured on Mochi. The response to the scale differs between models, so on another model the same measurements are repeated (Appendix~\ref{app:cogscale}); one pass with the raised scale per failure type is enough.

\section{Checks}
\label{sec:checks}

We tested five things that could undermine these results (full analyses in Appendix~\ref{app:checks}). First, failure cannot be predicted from the prompt text alone. Within rare composition, predictors built from the text (its complexity and a rating of how rare the scene is) are at chance on two of three models (AUROC .520 and .523, where .500 is chance), and so is the distance between $p$ and $\pd$ in the text encoder (.44 to .55), which compares the same two prompts as demand but never runs the video model. \emph{The text of a prompt does not tell us whether it will fail; that has to be read from the model.}

Second, demand improves on those predictors on all three models, though modestly (Table~\ref{tab:demand_main}b). \emph{It is best used, together with the failure type, to decide what to try after a failure, and not as a detector that says in advance whether a single prompt will fail.}

Third, demand is not specific to one model: measured on one model, it predicts untreated failures on another about as well as that model's own demand does. \emph{Demand measures something that the models share, whereas the conditions under which raising the scale helps differ from model to model (Appendix~\ref{app:cogscale}), so both should be measured on the model in use.}

Fourth, the link between demand and recovery is not a by-product of how one correction is built: in rare composition it holds for two differently built corrections and for the raised scale, which adds no direction. Fifth, the three suites' judges are comparable: a single judge asking one common question of every clip agrees with them to a moderate degree, and the disagreements fall mainly on borderline clips.

\section{Limitations and Future Work}
\label{app:limitations}
 
The predictive gains of demand are modest, so demand is a guide to what to try after a failure and not a detector of failures. The treated results are established on Mochi. On CogVideoX the raised scale responds differently (Appendix~\ref{app:cogscale}), so the conditions should be re-measured on each model; what carries over is the demand measurement. Repetitions with shared seeds, a re-seed control, and a sweep over five scales agree with the three outcomes reported here (Appendices~\ref{app:seeds} to~\ref{app:dose}). Every failure label comes from one judging model under fixed written instructions, which the agreement analysis of Sec.~\ref{sec:checks} addresses. Finally, the boundary we map is specific to corrections that act in the classifier-free guidance update; whether methods outside it obey the same demand threshold is the natural next question.

\section{Conclusion}

Raising the guidance scale is not a general remedy for a text-to-video model that ignores part of a prompt. It works under conditions that can be checked before generating again: on Mochi, when the prompt contains no negation and no event tied to a specific time, and when the demand score clears a fitted threshold (recovery about 65\% above it and about 39\% below it). The three outcomes are one behavior: the model generates the video of the default prompt instead of the request, and what differs is how it reacts to the part it drops. In practice, leave the scale alone for negations and temporal prompts, where the published directional and position-aware corrections apply (recovery .500 and .505), and otherwise read the demand score: raise the scale when it is high, and use the rare-composition correction when it is low. For evaluation, some prompts are beyond every guidance-based method we tested, so we recommend that a negative result be reported together with the demand of the prompts it failed on, which lets readers tell a weak method from a prompt that no such method repairs.

{
    \small
    \bibliographystyle{ieeenat_fullname}
    \bibliography{main}
}

\newpage
\onecolumn
\appendix

\section{Ethics Statement}
\label{app:ethics}
\begin{tcolorbox}[breakable, title=Ethics Statement]
This paper contributes a measurement, not a generation mechanism. It asks which text-to-video failures a guidance-based correction can reach, and answers by applying published corrections, unchanged, to frozen models.

Demand is intended as a diagnostic read before sampling. Where a failing prompt sits on the demand axis, together with its failure type, indicates whether the failure is likely to yield to a larger scale, to a directional correction, or to neither, so that compute is not spent asking guidance for what a model has not learned to give. The unified failure criterion and the recovery-against-demand map are meant to improve evaluation practice: they separate rendering quality from faithfulness, and they clarify what a negative result for a correction method does and does not show.

Any work that makes text-to-video generation more reliable can indirectly aid the production of deceptive or harmful video. This work adds no generative capability: no correction is proposed, no model is modified, and every treated generation uses a published method in its published configuration. The measurements read existing denoiser predictions and do not extend what any model can synthesize. The risk profile is therefore inherited from the underlying models, and the mitigations appropriate to them (content filtering, provenance marking, usage policies) remain applicable and necessary.

All prompts come from the three published evaluation suites, used unmodified and for evaluation only. They describe physically realizable natural and everyday scenes, were written without reference to identifiable individuals, and contain no offensive, sensitive, or privacy-implicating material. No human subjects were involved and no personal data was collected.

The unified judge, the per-suite judges, and the rarity rating $r$ query a proprietary multimodal model through a commercial API. The only content transmitted is the prompts and, for scoring, frames of videos generated in this study; no user data, private material, or third-party content is sent. Rubrics are fixed verbatim, decoding temperature is $0$, and the model version is recorded, so the protocol stays precisely specified even if the endpoint is later updated.

Nothing is trained or fine-tuned. Untreated measurement needs standard sampling plus demand replays, and because the first quarter of the steps suffices, the practical read costs a quarter of one generation's denoiser passes. Treated generation is confined to Mochi, apart from one scale-up transfer test on CogVideoX (Appendix~\ref{app:cogscale}). All experiments ran on standard research GPUs, and the absence of any training stage keeps the energy cost well below that of approaches that adapt a model.

A clip can look convincing while silently showing the default instead of the request. By making this measurable, before generation and from the model's own predictions, we hope to shift evaluation toward faithfulness, to make failure prediction and routing (regenerate, or apply the fitting correction) a standard step before generation, and to give correction methods a fair standard of evidence: a failure below the fitted demand threshold reflects the model's limit and not a weakness of the method.

Every pretrained model was retrieved from its official distribution and used under the terms its authors declare. Mochi~1~\cite{mochi} carries an Apache~2.0 license, which permits research use without restriction. The CogVideoX-5B checkpoint~\cite{yang2025cogvideox} is distributed under the CogVideoX License, which permits free use of the released models for academic research; our use is confined to that scope. HunyuanVideo~\cite{kong2024hunyuanvideo} is distributed under the Tencent Hunyuan Community License, which allows academic research within the licensed territory, and our experiments fall wholly inside that scope. The three corrections are used as published by their authors, and the CLIP-family models used by the embedding metrics permit the inference-only research use made of them. No weights are retrained, adapted, altered, or redistributed: each model is used strictly as a frozen sampler.
\end{tcolorbox}

\section{Extended Related Work}
\label{app:related}

\paragraph{Inference-time guidance control.}
Classifier-free guidance~\cite{ho2022classifier} has spawned a large family of training-free controls that reshape its update: semantic directions~\cite{brack2023sega, kang2023semantic}, attention-level methods~\cite{chefer2023attend, feng2022training, rassin2023linguistic}, composable formulations~\cite{liu2022compositional}, concept erasure and editing~\cite{gandikota2023erasing, gandikota2024unified}, constrained sampling~\cite{zampini2025training, zhang2025constrained}, and corrections to the update itself~\cite{chung2024cfg++, si2024freeu, bansal2023universal, xie2025dymo}. Known problems of the scale are also increasingly documented: one scale value does not suit all regions of an image equally~\cite{shen2024rethinking}, and raising it globally also strengthens attributes that the user did not ask for~\cite{xia2025factored}. What this literature does not tell us is which failures cannot be repaired by any method of this kind, that is, by any method that only changes the guidance update. To study this, we take published corrections and apply them exactly as published, as measuring instruments (the criteria for choosing them are in Sec.~\ref{sec:related}), and we add the raised scale as a fourth method (Appendix~\ref{app:scope}).

\paragraph{Three failure types.}
\emph{Rare composition} builds on the compositionality deficits of vision-language systems~\cite{yuksekgonul2022and, thrush2022winoground, hsieh2023sugarcrepe}; generation-side remedies range from LLM-guided rewriting~\cite{park2025rare} to counterfactual guidance~\cite{kang2026dcr}.
\emph{Negation} insensitivity is documented across vision-language models~\cite{alhamoud2025vision, zhang2025negvqa}, with trained remedies~\cite{singh2024learn} and training-free ones~\cite{xiao2026notjustwhatsthere, kang2026negate}. Mechanistically, \cite{zhou2026language} show that language models internally \emph{construct} the negated alternative and fail through shortcut attention. This is evidence from another modality that a negation failure can be the suppression of a result the model does represent, and it is the distinction that our demand measurements draw for video.
\emph{Temporal} emergence connects to inference-time temporal control~\cite{wu2025mind, cai2025ditctrl, lo2025prompt, chen2026prompt, schiber2026tempocontrol, kang2025text2story}, to temporally self-contradicting prompts~\cite{huberman2026image}, and to the decoupling correction we adopt~\cite{kang2026tpd}.
The three lines of work report the same qualitative diagnosis. This paper tests it by measuring the three together.

\paragraph{Model-internal signals as measurement.}
Concept Removal Guidance estimates the presence of a concept at each step from differences of noise predictions~\cite{choi2026concept}. That validates signals of exactly our form, although there the signal steers sampling and ours is never fed back.
Closest to our axis, \cite{li2026initialization} show that a guidance-potential distance over the initial noise governs mode collapse in image generation. We share the premise that distributional distance is the operative variable and differ in the object (video), the unit (a prompt-model pair over the early steps), and the question (which failures any guidance-based correction can reach).
Learned identification of where a method can act appears in~\cite{meshchaninov2026guided}; ours is training-free and available before generation.
Phase-dependent editability along the noise scale~\cite{lu2026semantic} motivates our finding that failure is settled during structure formation.
Locate-Then-Correct~\cite{yeo2026locate} describes a two-stage approach: first find where a model goes wrong, then correct it. Our paper deliberately carries out only the first stage. We identify which failures a correction can repair, and we do not propose a correction of our own.

\paragraph{CLIP similarity.}
Our embedding-side measure (Eq.~\ref{eq:realized}) heeds two cautions about CLIP-family similarity: raw similarity misleads about where a referent \emph{is}~\cite{lee2026clip}, and negated content is systematically misread~\cite{alhamoud2025vision}. Measuring a clip's position between two fixed references, $p$ and $\pd$, addresses both and makes absence measurable (Appendix~\ref{subsec:realized}).

\section{Notation and Formal Definitions}
\label{app:formal}

The main text describes several analyses in words. This section states them.

\paragraph{Notation.}
Let $x_t \in \mathbb{R}^{F \times H \times W \times C}$ be the latent at timestep $t$ with $F$ frames, and $\epsilon_\theta(x_t,t,c)$ the frozen denoiser of model $\theta \in \Theta$ under conditioning $c=\Phi_\theta(y)$ for text encoder $\Phi_\theta$. We write $\epsilon_{\text{text}} := \epsilon_\theta(x_t,t,\Phi_\theta(p))$, $\epsilon^{\circ} := \epsilon_\theta(x_t,t,\Phi_\theta(\pd))$, and $\epsilon_{\text{uncond}} := \epsilon_\theta(x_t,t,\varnothing)$, all on the same latent. On HunyuanVideo, whose distilled sampling does not compute an unconditional branch, $\epsilon_{\text{uncond}}$ is obtained by one additional pass of the frozen denoiser with the empty prompt on the same latent; the sampling run itself is unchanged. The guidance update~\cite{ho2022classifier} is $\delta_{\text{ref}} = w(\epsilon_{\text{text}} - \epsilon_{\text{uncond}})$ with scale $w>1$, and the difference that demand measures is $d_t = \epsilon_{\text{text}} - \epsilon^{\circ}$. We suppress $\theta$ except where the dependence on the model matters. $\mathcal{V}$ denotes a failure rate, $v \in \{0,1\}$ the failure label of one clip, and $D^{*}$ a fitted demand threshold.

\subsection{The Default Prompt as an Operator}

The construction of Sec.~\ref{subsec:default_operator} is an operator on text,
\begin{equation}
\mathcal{O} : p \;\longmapsto\; \pd,
\label{eq:default_operator}
\end{equation}
which deletes the improbable specification and preserves the surrounding scene: it relaxes the rare factor, deletes the negation and its scope, or cuts the late event. $\mathcal{O}$ acts on text alone, so one $\pd$ serves every model.

Each of the three corrections also builds a reference from the denoiser, but for \emph{correcting}, and each builds it differently: a forbidden span~\cite{kang2026negate}, a common counterpart~\cite{kang2026dcr}, a truncated early scene~\cite{kang2026tpd}. Our operator is a reference for \emph{measuring}: one construction, applied identically in all three failure types, and never fed back into sampling. For temporal prompts it coincides textually with the reference of~\cite{kang2026tpd}; Appendix~\ref{app:identification} explains how measurement and correction are kept apart there.

\subsection{Demand over the First $K$ Steps and Transfer Across Models}

Demand (Eq.~\ref{eq:demand}) averages the ratio $\|d_t\|_2 / \|\epsilon_{\text{text}} - \epsilon_{\text{uncond}}\|_2$ over a set of recorded steps of an untreated run. $\dem^{(K)}$ averages it over only the first $K$ recorded steps, the ones with the most noise. The main text uses $K = T/4$ throughout and calls it first-quarter demand.

Because demand depends on the model, it supports a test that no quantity computed from the prompt text can: predicting failure on one model from demand measured on another. For an ordered pair of models $(\theta,\theta')$ we fit
\begin{equation}
\Pr\!\left[v = 1;\, \theta'\right] \;=\; \sigma\!\left(\beta_0 + \beta_1\, \dem(p;\theta) + \boldsymbol{\beta}_f^{\top}\mathbf{f}\right),
\label{eq:transfer}
\end{equation}
where $v$ is the failure label of the untreated clip from $\theta'$ and $\mathbf{f}$ are failure-type indicators, and we report the matrix of AUROCs $A_{\theta\theta'}$ (Table~\ref{tab:transfer}). In this matrix a diagonal entry uses demand and failure labels from the same model, and an off-diagonal entry uses demand from one model to predict the failures of a different model. If the off-diagonal entries are nearly as large as the diagonal ones, then what demand measures about a prompt is largely the same on every model. The test also answers a possible objection, namely that demand might predict a correction's success only because demand and the correction are computed from the same quantities. That cannot be the cause here: demand is computed with one model and the outcome with another, so no formula connects the two.

\subsection{An Embedding-Side Measure and the Unified Failure Judgment}
\label{subsec:realized}

The embedding-side measure asks how far a clip moved in CLIP space. For a clip of $T$ sampled frames $\{I_j\}$,
\begin{equation}
\mathcal{D}_{\text{real}}(x_0; p)
= \frac{1}{T}\sum_{j=1}^{T}
\Big[\cos\!\big(E_{\text{img}}(I_j), E_{\text{text}}(p)\big)
- \cos\!\big(E_{\text{img}}(I_j), E_{\text{text}}(\pd)\big)\Big],
\label{eq:realized}
\end{equation}
where $E_{\text{img}}$ and $E_{\text{text}}$ are the image and text encoders of a frozen CLIP model. Positive values mean the clip resembles the request more than the default; values near zero or below mean the clip shows the default. The difference form makes absence measurable: a concept that must \emph{not} appear cannot raise similarity to $p$, but its removal lowers similarity to $\pd$, so negation and presence sit on one scale. The second term reduces to the suites' own metrics (CLIP-attr, CLIP-neg, CLIP-early) under the corresponding choice of $\pd$, so those are special cases. For temporal prompts we also evaluate Eq.~\eqref{eq:realized} on the late frames only. The embedding-side effect of a method $\mathcal{M}$ is $\Delta(\mathcal{M}; p) = \mathcal{D}_{\text{real}}(x_0^{\mathcal{M}}; p) - \mathcal{D}_{\text{real}}(x_0; p)$.

\label{subsec:collapse}The unified judgment asks one binary question in all failure types. A multimodal judge receives $p$, $\pd$, uniformly sampled frames in order, and, for prompts that place an event in time, the frame partition. It returns
\begin{equation}
v = \mathbf{1}\!\left[\text{the clip shows } \pd \text{ rather than } p\right],
\qquad
\mathcal{V} = \frac{1}{N}\sum_{i=1}^{N} v_i .
\label{eq:collapse}
\end{equation}
Each suite's original judge asks its own version of this question. The negation judge asks whether the forbidden content still appears. The rare-composition judge asks whether the clip shows the common version of the scene instead of the rare one. The temporal judge asks whether the late event failed to arrive. In all three cases a `yes' means that the clip shows $\pd$ rather than $p$, which is the single question of Eq.~\eqref{eq:collapse}. The unified judge is given the same written instructions for every clip, runs at temperature $0$ so that its output does not vary between runs, and scores every clip with none excluded. Appendix~\ref{app:kappa} measures how well it agrees with each suite's original judge. All analyses in the paper use the original judges' labels; the unified judge serves only to check that those labels are comparable across failure types.

\subsection{Text-Only Predictors}
\label{subsec:predictors}

If demand predicts failure, the alternatives must get a fair chance to predict it instead. For each prompt we compute predictors that never consult the denoiser:
\begin{itemize}
\setlength{\itemsep}{0pt}
\item \emph{complexity} $\mathbf{c}$: token length, content-word count, number of distinct entities, modifier count, and syntactic depth of the clause that carries the difficult specification;
\item \emph{rarity} $r(p)$: how rare the scene is in the world, rated by an external language model under a fixed rubric at temperature $0$;
\item \emph{encoder distance} $u(p) = 1 - \cos\big(\bar{\Phi}(p), \bar{\Phi}(\pd)\big)$, from the pooled embedding of the same frozen text encoder that conditions generation.
\end{itemize}
Failure is modeled as
\begin{equation}
\Pr[v = 1] \;=\; \sigma\!\left(\beta_0 + \beta_1 \dem + \boldsymbol{\beta}_c^{\top}\mathbf{c} + \beta_u\, u + \beta_r\, r + \boldsymbol{\beta}_f^{\top}\mathbf{f}\right),
\label{eq:regression}
\end{equation}
with failure-type fixed effects $\mathbf{f}$ that absorb differences between suites, so a significant $\beta_1$ reflects variation within failure types. We report the AUROC the model gains when the demand term is added ($\Delta$AUROC). AUROC has its elementary reading: draw one failing and one succeeding prompt at random, and AUROC is the probability that the score ranks them correctly; chance is $0.500$. The comparison against $u$ is the sharpest one. $u$ shares our text encoder and our default prompt and differs only in never consulting the denoiser, so any gain of demand over a set of predictors that includes $u$ isolates what the denoiser contributes.

\subsection{Recovery Along the Demand Axis}

Within each failure type we fit a segmented (two-slope) regression of recovery on demand, with bootstrap confidence intervals on the breakpoint $D^{*}$. Each failure type receives its own fit. Whether the bend is sharp or gradual is an empirical property that we report and do not assume (Table~\ref{tab:segmented}).

\section{Test Suites and Scale}
\label{app:suites}

The three suites are those of their source papers~\cite{kang2026negate, kang2026dcr, kang2026tpd}, used unmodified. Each has 8 categories $\times$ 50 prompts $=$ 400 prompts, for 1{,}200 unique prompts in all. With three models, the untreated set holds 3{,}600 labeled clips, one per prompt-model pair. Seeds are not shared across conditions, so a paired comparison contains seed noise in addition to the effect of the method; Appendix~\ref{app:seeds} measures how large that noise is. The methods add $8 \times 400 = 3{,}200$ judged clips: six cells in which a correction is applied outside its own failure type, and two scale-up cells. The check of the default prompt adds 24 clips.

Category definitions follow the source suites verbatim; Table~\ref{tab:categories} lists the codes and what each failure type's categories vary. The excluded content in the negation suite ranges from categorical (an object class) to graded (an attribute level).

\begin{table}[ht]
\centering
\small
\begin{tabular}{@{}lp{0.36\linewidth}p{0.42\linewidth}@{}}
\toprule
Failure type & Category codes & What the categories vary \\
\midrule
Negation~\cite{kang2026negate} & AOC, LEN, INA, MNC, SFN, NMI, DNS, SND & \emph{what the prompt excludes, and how}: Absent Object Consistency (AOC, \emph{a highway with no vehicles}); Late Emergence Negation (LEN, \emph{no performers appearing}); Implicit Natural-Only Attribute (INA, exclusion implied by \emph{only}, with no negation word); Multi-Negation Composition (MNC, \emph{no desks, no chairs, and no books}); Structural Functional Negation (SFN, \emph{holding a tool but not using it}); Non-Inversion Mitigation (NMI, graded, \emph{not aggressive}); Double Negation Sensitivity (DNS, \emph{not unlit}); and Scoped Negation Disambiguation (SND, \emph{chasing a ball that is not moving}) \\
\addlinespace
Rare composition~\cite{kang2026dcr} & ENV, TEMP, OBJ, ATTR, SCALE, CTX, MAT, DENS & \emph{which binding is made improbable}: Environment Recomposition (ENV, \emph{a snowy beach}); Temporal Misalignment (TEMP, \emph{a rainbow at night}); Object Relocation (OBJ, \emph{a penguin in a sandy desert}); Attribute Rebinding (ATTR, \emph{a glowing wooden chair}); Scale Shift (SCALE, \emph{a giant apple next to a small house}); Contextual Relocation (CTX, \emph{a classroom inside a forest}); Material and State Conflict (MAT, \emph{a frozen waterfall in summer}); and Composition Density Variation (DENS, \emph{a nearly empty street at noon}) \\
\addlinespace
Temporal~\cite{kang2026tpd} & ENTRY, SCENE, ONSET, STATE, CAUSAL, WEATHER\_NIGHT, ACTIVITY, STRUCTURAL & \emph{what kind of change must emerge}: entity entry, scene transition, onset of a process, state change, a causal consequence, a weather-to-night shift, an activity change, and structural transformation \\
\bottomrule
\end{tabular}
\caption{\textbf{Categories of the three suites.} Eight categories of 50 prompts each per failure type. Category names follow the papers that released the suites; example prompts in italics.}
\label{tab:categories}
\end{table}

One pair of categories is instructive. Negation LEN (Late Emergence Negation) and temporal ENTRY both concern the arrival of content in a clip, but LEN forbids it and ENTRY requires it. The two suites thus constrain opposite directions along the same time axis, and the unified judgment (Eq.~\ref{eq:collapse}) scores both with the same question and opposite polarity. Demand separates the pair as the main text predicts: failing LEN prompts score high (a strong, mis-aimed reaction) and failing ENTRY prompts score low (a weak one), matching the directions of Table~\ref{tab:demand_sign}.

\section{The Unified Judgment and Its Agreement with the Original Judges}
\label{app:kappa}

The agreement sample is stratified with one clip per cell over evaluation files $\times$ categories $\times$ label values: 13 files $\times$ 8 categories $\times$ 2 labels $= 208$ cells. The 13 files are the nine untreated files (three models $\times$ three failure types), the three files of each correction applied to its own failure type on Mochi, and the temporal scale-up file. Seven cells contain no clip with the required label (for example, negation INA has no untreated failure on any model), which gives $n=201$. Each judged clip is represented by 8 uniformly sampled frames, under the fixed rubric at temperature 0. Per-category sample sizes (4 to 10) support raw agreement but not a per-category $\kappa$, so we report $\kappa$ overall and per failure type only.

\begin{figure}[ht]
    \centering
    \includegraphics[width=0.85\linewidth]{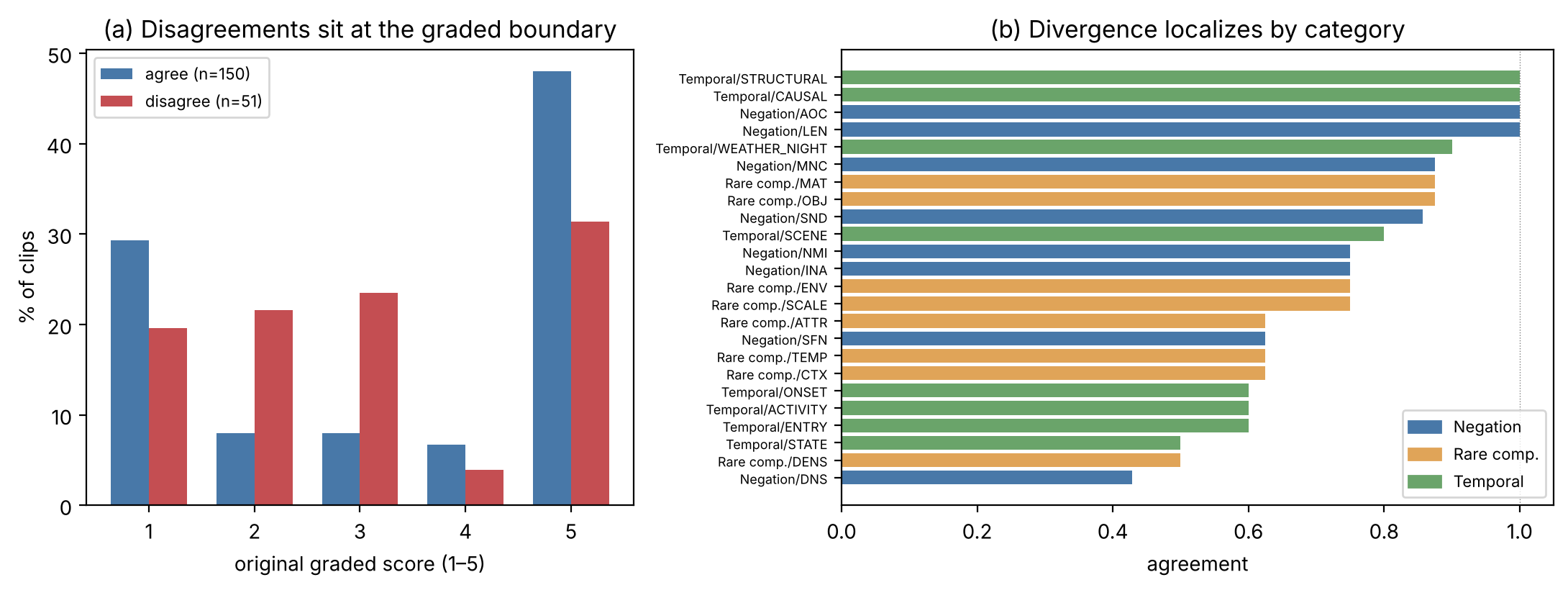}
    \caption{\textbf{Where the two judging protocols part.} (a)~Original graded scores (1 to 5) of clips on which the protocols agree (blue) and disagree (red). Agreements concentrate at the extremes of the scale; disagreements carry intermediate scores at more than twice the rate ($49\%$ against $23\%$; Mann-Whitney $p=1.1\times10^{-4}$). (b)~Agreement per category, sorted; labels give failure type/category. Agreement is perfect where failure destroys or transforms the scene (temporal CAUSAL and STRUCTURAL) or where the forbidden content is categorical (negation AOC, LEN). Divergence is confined to scene-persistence categories, in which even a successful clip contains the default as a sub-description, and to graded-attribute categories, where the binary cut divides a continuum.}
    \label{fig:kappa}
\end{figure}

Agreement is $\kappa = 0.49$ overall and $0.41$ to $0.59$ per failure type (Table~\ref{tab:kappa}). Figure~\ref{fig:kappa} shows where the disagreement comes from. First, the two judgments disagree mostly on borderline clips. The original judges give each clip a graded score from 1 to 5 before deciding between failure and success, and a middle score means that the clip only partly follows the prompt. Clips on which the two judgments disagree have a middle score more than twice as often as clips on which they agree (49\% against 23\%). Second, the disagreement is limited to certain categories. Where failure is all-or-nothing, for example when an object that should be absent is either in the clip or not, the two judgments agree perfectly. They diverge in two kinds of category: those in which the judged property comes in degrees (for example \emph{not aggressive}), and those in which a correct clip necessarily also shows the default scene (a clip in which the wave arrives still shows a sandcastle on a beach first), so that the question of whether the clip shows $\pd$ is harder to answer. Both judgments thus measure the same thing, and they differ only in where they place the line between failure and success for a clip that lies in between.

The validation extends to the treated clips. The eight sets of treated clips in Table~\ref{tab:grid} were scored by the \emph{same} per-suite judges, with rubrics, frame protocol, judge model, and decoding parameters identical to those validated here. The agreement analysis therefore transfers at the level of the judge.

\begin{table}[ht]
\centering
\small
\begin{tabular}{lcccc}
\toprule
 & $n$ & Agreement & Cohen's $\kappa$ & 95\% CI \\
\midrule
Overall            & 201 & 0.746 & 0.489 & $[0.368,\, 0.605]$ \\
Negation           & 58  & 0.793 & 0.593 & $[0.394,\, 0.791]$ \\
Rare composition   & 64  & 0.703 & 0.406 & $[0.206,\, 0.600]$ \\
Temporal           & 79  & 0.747 & 0.496 & $[0.318,\, 0.665]$ \\
\bottomrule
\end{tabular}
\caption{\textbf{Agreement between the unified judgment and the original per-suite judgments.} Cohen's $\kappa$ with bootstrap 95\% CIs, on a stratified sample that covers every model, failure type, and label value.}
\label{tab:kappa}
\end{table}

\section{Untreated Failure by Category and Model}
\label{app:atlas}

\begin{figure}[ht]
    \centering
    \includegraphics[width=\linewidth]{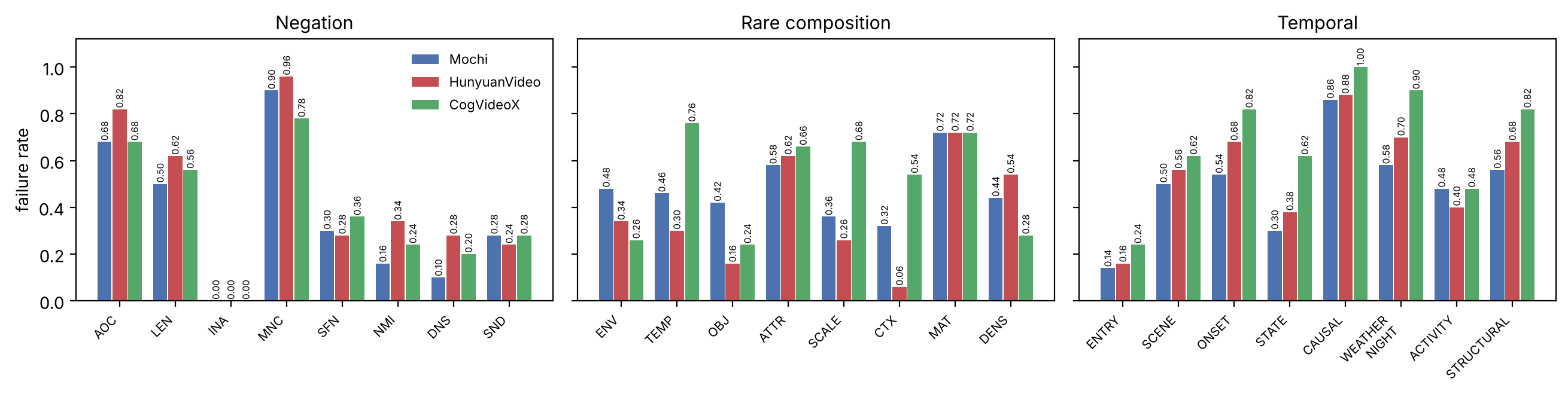}
    \caption{\textbf{Untreated failure rate by failure type, category, and model.} Cells span the full interval $[0.00, 1.00]$.}
    \label{fig:atlas}
\end{figure}

Table~\ref{tab:atlas} gives the failure rate of each failure type on each model, and how strongly two models agree on \emph{which prompts} fail. Figure~\ref{fig:atlas} resolves the failure rate by category.

Models agree on which categories are hard more than on which prompts are hard. The ordering of categories by difficulty is broadly preserved across models, while levels shift and individual categories reverse (Temporal Misalignment, rare-composition TEMP: $0.46$ on Mochi against $0.76$ on CogVideoX; Contextual Relocation, CTX: $0.32$ against $0.06$ on HunyuanVideo).
At the level of single prompts, failure is substantially shared for negation and temporal prompts ($\phi$ about .37 to .50) and only weakly shared in rare composition ($\phi$ about .16 to .26). This is the first indication that failure is a property of the prompt-model pair and not of the prompt alone.

The rates also span the whole range. Cells run from $0.00$ to $1.00$: some categories fail on every model almost always (temporal CAUSAL on CogVideoX at $1.00$), and in one no model fails at all (negation INA at $0.00$ on all three). This spread is what a predictive account needs. The number of models that fail a given prompt is likewise spread over its whole range in every failure type, which anticipates the transfer analysis of the main text.

\begin{table}[ht]
\centering
\small
\begin{tabular}{lccc@{\hspace{1.6em}}ccc}
\toprule
& \multicolumn{3}{c}{Failure rate $\mathcal{V}$} & \multicolumn{3}{c}{Prompt-level agreement $\phi$} \\
\cmidrule(r){2-4}\cmidrule(l){5-7}
Failure type & Mochi & Hunyuan & CogVX & M/H & M/C & H/C \\
\midrule
Negation           & 0.365 & 0.442 & 0.388 & 0.496 & 0.474 & 0.480 \\
Rare composition   & 0.472 & 0.375 & 0.518 & 0.260 & 0.162 & 0.190 \\
Temporal           & 0.495 & 0.555 & 0.688 & 0.383 & 0.365 & 0.384 \\
\bottomrule
\end{tabular}
\caption{\textbf{Untreated failure and how far models share it.} Left: failure rate per failure type under each model's default sampling settings. Right: $\phi$ correlation of the per-prompt failure label between pairs of models.}
\label{tab:atlas}
\end{table}

\section{Recovery Against the Number of Models That Fail}
\label{app:difficulty}

\begin{figure}[ht]
    \centering
    \includegraphics[width=0.6\linewidth]{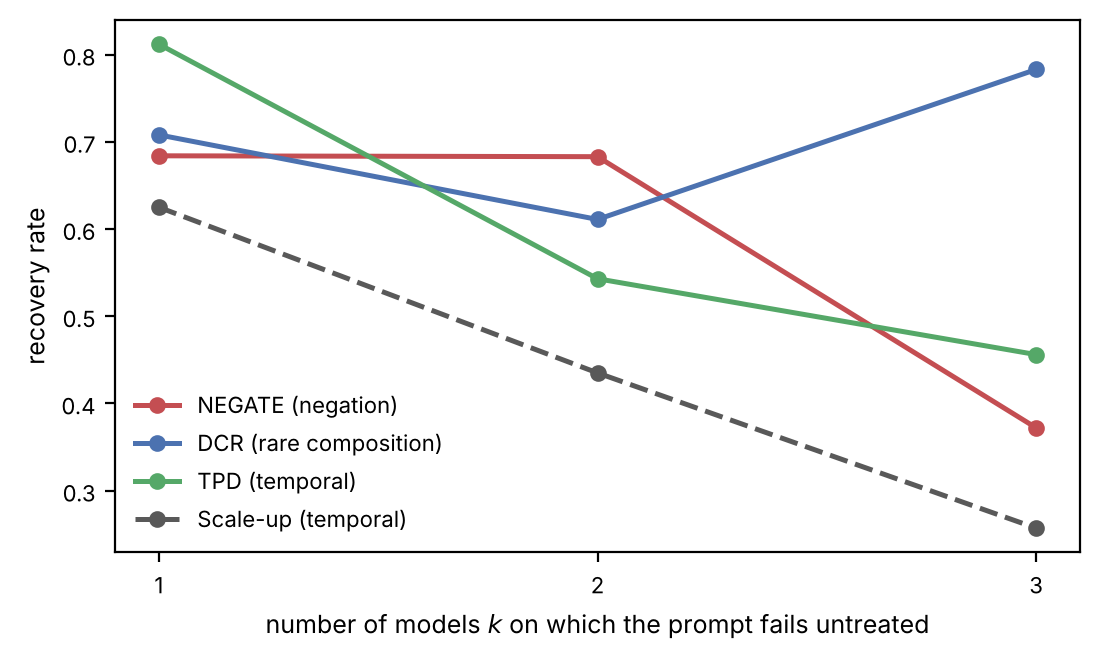}
    \caption{\textbf{Recovery against the number $k$ of models on which a prompt fails.} For prompts that fail on Mochi, the probability that the correction designed for that failure type repairs them. NEGATE (negation) and TPD (temporal): recovery is lowest for prompts that fail on all three models ($0.68 \to 0.68 \to 0.37$; $0.81 \to 0.54 \to 0.46$). Scale-up on temporal prompts (dashed) lies below TPD at every $k$ ($0.63 \to 0.44 \to 0.26$). DCR (rare composition): no ordering ($0.71 \to 0.61 \to 0.78$), because models barely agree on which rare-composition prompts fail ($\phi \approx 0.2$).}
    \label{fig:recovery_k}
\end{figure}

Before introducing demand, we tried a simpler measure of how hard a prompt is: the number $k \in \{1,2,3\}$ of models on which it fails untreated. Figure~\ref{fig:recovery_k} plots, for prompts that fail on Mochi, how often the correction designed for that failure type repairs them, as a function of $k$. Table~\ref{tab:treatment_label} gives the corresponding overall rates.

For failure types in which the three models tend to fail on the same prompts, $k$ is a reasonable measure of how hard a prompt is: a prompt that fails on all three models is harder than one that fails on only one, and harder prompts should be harder to repair. This is what we observe. For negation and temporal prompts, recovery does not rise with $k$ and is lowest at $k{=}3$ ($0.68 \to 0.68 \to 0.37$ for negation; $0.81 \to 0.54 \to 0.46$ for temporal). Prompts that defeat all three models resist correction at roughly twice the rate of prompts that defeat only one. Scale-up on temporal prompts (dashed) lies below TPD at every $k$ ($0.63 \to 0.44 \to 0.26$); weighted by bin sizes, these values aggregate to its overall recovery of .328 (Table~\ref{tab:treatment_label}).

Where models do not share failures, $k$ is noise. In rare composition no ordering emerges ($0.71 \to 0.61 \to 0.78$). With prompt-level agreement of only $\phi \approx 0.2$, $k$ counts sampling noise and not difficulty.

A count of labels across models therefore cannot say which failures are treatable in the failure type where that is most needed. A usable measure must be continuous, defined per prompt and per model, and meaningful where models disagree. Demand is defined to meet these three requirements.

\begin{table}[ht]
\centering
\small
\begin{tabular}{llcccc}
\toprule
Failure type & Method & Recovery & Induction & Net change in $\mathcal{V}$ & McNemar $p$ \\
\midrule
Negation           & NEGATE (own type) & 0.500 & 0.075 & $-0.135$ & $1.2\times10^{-8}$ \\
Rare composition   & DCR (own type) & 0.698 & 0.289 & $-0.178$ & $3.5\times10^{-7}$ \\
Temporal           & TPD (own type) & 0.505 & 0.153 & $-0.172$ & $1.2\times10^{-9}$ \\
Temporal           & Scale-up & 0.328 & 0.386 & $+0.032$ & $0.32$ \\
\bottomrule
\end{tabular}
\caption{\textbf{Each correction on its own failure type, and scale-up on temporal prompts} (Mochi). Recovery: fraction of untreated failures that succeed under the method. Induction: fraction of untreated successes that fail under it. Net change: change in the failure rate. McNemar $p$: exact binomial test on the pairs that changed. Rows describe what guidance can do per failure type and are not to be compared with one another.}
\label{tab:treatment_label}
\end{table}

\section{Protocol for Applying Corrections Outside Their Own Failure Type}
\label{app:grid}

Each correction needs a second prompt in addition to the user's prompt. This reference prompt describes what the generation should move away from: the forbidden content for NEGATE, the common version of the scene for DCR, and the scene before the event for TPD. When a correction is applied to the failure type it was designed for, its own procedure produces the reference prompt. When we apply it to another failure type, that procedure does not apply, and we supply the default prompt $\pd$ of the prompt at hand as the reference. All other settings of the correction keep their published values (Table~\ref{tab:grid_protocol}). The raised scale needs no reference prompt; scale-up changes only the guidance scale, to the value published with the temporal suite. Two implementation details complete the protocol. The published implementation of NEGATE derives the reference with a linguistic extractor that is defined only for negation syntax, so outside negation we pass $\pd$ to the reference slot directly, through an added code path that leaves the published pipeline byte-identical when unused. The decomposition inputs of TPD have no referent outside temporal prompts, so we set them to their limiting values: $\gamma = 0$ (the event window is the whole clip) and $(\rho_{\text{early}}, \rho_{\text{late}}) = (1, 0)$, at which the adaptive coefficient equals its published base value $\alpha_0$. The symmetric choice $(0.5, 0.5)$ would silently disable the correction, since $\alpha = \alpha_0 \max(0, \rho_{\text{early}} - \rho_{\text{late}})$.

Each combination of a method and a failure type is one cell of Table~\ref{tab:grid}. The clips of a cell are labeled by the judge of the failure type that the prompts belong to (the \emph{target} failure type), not by the judge of the suite the method came from. For example, when NEGATE is applied to rare-composition prompts, the rare-composition judge decides whether each clip is a failure. The reason is that the prompt determines what counts as success, whichever method is applied. Each cell is then compared with the untreated clips of the same prompts, labeled by the same judge. We do not compare two methods with each other directly, with one exception: the decision rule has to choose between DCR and the raised scale on rare composition, so Sec.~\ref{par:rule} compares them, on net recovery (Appendix~\ref{app:reseed}). That comparison stays within one failure type and one judge, and both cells are paired, prompt by prompt, with the same untreated run. The remaining four cells of Table~\ref{tab:grid} (each correction on its own failure type, and scale-up on temporal prompts) use the clips released with the source papers' evaluation runs; their recovery and induction are recomputed here by pairing each treated clip, prompt by prompt, with the same untreated Mochi run that serves every other cell, so all twelve cells share one untreated reference.

\begin{table}[ht]
\centering
\small
\begin{tabular}{llll}
\toprule
Method & Applied to & Slot that receives $\pd$ & Scored by \\
\midrule
NEGATE & rare composition, temporal & forbidden-content slot & target failure type's judge, vs.\ untreated \\
DCR    & negation, temporal   & common-counterpart slot  & target failure type's judge, vs.\ untreated \\
TPD    & negation, rare composition & early-scene slot ($\gamma{=}0$) & target failure type's judge, vs.\ untreated \\
Scale-up & negation, rare composition & none (scale only) & target failure type's judge, vs.\ untreated \\
\bottomrule
\end{tabular}
\caption{\textbf{The eight cells generated for this study.} Each correction runs on Mochi in its published configuration, with its reference slot filled by the target failure type's $\pd$. The three cells in which a correction meets its own failure type and the temporal scale-up cell complete the grid of Table~\ref{tab:grid}.}
\label{tab:grid_protocol}
\end{table}

\paragraph{Selection of qualitative examples.}
For every qualitative figure, candidates were enumerated from the judged clips by a fixed condition, and the displayed example was chosen among the candidates for visual clarity only. The conditions are:
\begin{itemize}
\setlength{\itemsep}{0pt}
\item Fig.~\ref{fig:default}: the $\pd$ clip is judged to show the default, and the untreated clip for the source prompt is judged a failure;
\item Fig.~\ref{fig:rx_negation}: untreated fails, scale-up fails, NEGATE succeeds;
\item Fig.~\ref{fig:rx_rarecomp}: untreated fails, scale-up succeeds, DCR succeeds, demand above the failure type's threshold;
\item Fig.~\ref{fig:rx_temporal}: untreated fails, scale-up fails, TPD succeeds;
\item Fig.~\ref{fig:switch}: untreated fails, all four methods fail, at least one other model succeeds untreated.
\end{itemize}

\section{Choice of Corrections and of Mochi}
\label{app:scope}

The three criteria of Sec.~\ref{sec:related} (the method acts purely within the classifier-free guidance update, it has a validated configuration on Mochi, and it ships an evaluation suite for one failure type) admit NEGATE, DCR, and TPD. We know of no other published correction that satisfies all three.

Ablated variants of the corrections are excluded by design. The corrections enter as fixed, published probes; an ablation asks why a method works, and this paper asks what the class of methods can reach. Attention surgery, retraining, and classifier-guided controls act outside the guidance update and are out of scope.

All treated generations with the three corrections run on Mochi, because they publish their validated configurations there; the raised scale is additionally run on CogVideoX as a transfer test (Appendix~\ref{app:cogscale}). Apart from that test, the other two models contribute untreated clips, demand measurements, and targets for the transfer analysis, which is what the claims across models require of them.

\section{Scale-up on a Second Model}
\label{app:cogscale}

The scale-up arm was repeated on CogVideoX for the negation and rare-composition suites. These runs postdate the main analyses and are reported here as a test of whether the conditions transfer; they are not folded into the main tables.

The second model has to be CogVideoX. Raising the scale means increasing the weight $w$ of standard classifier-free guidance, so the test requires a model on which $w$ exists as a free parameter at inference time. Mochi and CogVideoX expose it. HunyuanVideo does not: its guidance is distilled into the model during training, and what its interface calls a guidance scale is a learned embedding fed to the network, not the weight $w$. Raising that embedding is a different operation from raising $w$, since it gives the model an input outside the range it was trained on, and switching HunyuanVideo to its optional true-CFG mode would change the sampling procedure itself, so the treated and untreated clips of a prompt would then differ in more than the scale. On HunyuanVideo the operation this appendix tests is therefore not defined, and running it anyway would measure something else. CogVideoX uses the same standard two-branch guidance as Mochi, so the raised scale means the same thing on both.

CogVideoX has no published scale-up baseline, so the guidance scale was raised from the untreated 7.0 to 12.0; the Mochi cell uses the configuration published with the temporal suite (4.5 to 9.0), so the relative increase here (factor 1.7) is smaller than the one Mochi tolerated (factor 2.0). Scoring follows Appendix~\ref{app:grid}: each suite's original judge, each cell against the untreated CogVideoX run, prompt by prompt.

\begin{table}[ht]
\centering
\small
\begin{tabular}{lcccccc}
\toprule
Failure type & Untreated & Scale-up & Recovery & Induction & Net & McNemar $p$ \\
\midrule
Negation         & .388 & .255 & .477 & .086 & $-.133$ & $4.2\times10^{-8}$ \\
Rare composition & .518 & .718 & .111 & .534 & $+.200$ & $2.8\times10^{-13}$ \\
\bottomrule
\end{tabular}
\caption{\textbf{Scale-up on CogVideoX} (failure rates; recovery and induction as in Table~\ref{tab:grid}). The rare-composition effect runs opposite to Mochi's ($-.175$ there), and negation, which the raise does not help on Mochi (net within seed noise; Appendix~\ref{app:seeds}), is clearly helped here.}
\label{tab:cogscale}
\end{table}

The response of the raised scale differs by model. On Mochi it lowers the rare-composition failure rate, and on negation it repairs few failures, with a net change whose sign the seeds do not settle (Appendices~\ref{app:seeds} and~\ref{app:dose}). On CogVideoX the rare-composition effect inverts into harm, and negation, which no raise helps on Mochi, is clearly helped (Table~\ref{tab:cogscale}). The conditions of Fig.~\ref{fig:rule} are therefore measured outcomes on Mochi, not constants of a failure type.

A third of the induced failures are breakdowns of generation. The 103 rare-composition prompts that succeeded untreated and failed under scale-up were inspected clip by clip. Thirty-three (32\%) show no coherent scene at all: white or near-uniform frames, abstract textures, in one case rendered text, with nothing for a judge to assess. CogVideoX degenerates at a relative scale increase smaller than Mochi tolerated, so this component is consistent with a narrower usable guidance range rather than content. The remaining seventy (68\%) are coherent videos that converged to the common counterpart, and the judge's written evidence names the counterpart in 41 of the 70 (a typical sunrise without the requested stars, a clear sky without the requested moon, dry sand without the requested frost). Excluding every breakdown still leaves an induction of .363 against Mochi's .152, so the inversion is not an artifact of the breakdowns: on this model, amplification itself drives weakly registered requests toward the default.

The demand ordering survives for negation. Among the 155 untreated negation failures, recovery under scale-up falls as first-quarter demand (measured on CogVideoX) rises: $r=-.260$, $p=.0011$, with tercile recoveries $.64/.47/.33$ from low to high demand. The same downward slope appears on Mochi, where the re-seed control shows that re-sampling alone falls the same way, because high-demand negation failures are persistent (Appendix~\ref{app:reseed}). CogVideoX has no such control, so this slope too may belong to the failures rather than to the method. In rare composition the relation is flat ($r=-.004$; terciles $.13/.07/.13$), consistent with Table~\ref{tab:demand_sign}, where CogVideoX's rare-composition demand-failure correlation ($-.070$) was already the weakest of the three models.

Taken together, these runs show that the response to a method, like failure itself, is a property of the prompt-model pair. The conditions of Fig.~\ref{fig:rule} are validated on Mochi; on another model they should be re-measured, not assumed. What transfers is the measurement and, for negation, the ordering of its failures by demand.

\section{Seed Robustness of the Measurements}
\label{app:seeds}

The main protocol generates one clip per prompt, and the untreated and the treated clip of a prompt use different random seeds. A result could therefore reflect the luck of one seed. To find out how much, we repeated part of the experiment three times with controlled seeds. Three things came out. First, demand hardly depends on the seed. Second, the finding that the raised scale has no net effect on temporal prompts is stable. Third, for negation every run agrees on what the rule of Fig.~\ref{fig:rule} uses, namely that the raised scale repairs few failures; the net change in the failure rate is small in every run, of the size of the seed noise, and its sign varies between replicates.

\paragraph{Design.}
We repeated the two scale-up cells whose statistics are borderline in Table~\ref{tab:grid}: negation ($+.042$ at $p=.05$) and temporal ($p=.32$). The first 20 prompts of each of the eight categories of the two suites were fixed before generation, which gives 160 prompts per failure type. Each prompt was generated on Mochi at the untreated scale (4.5) and at the raised scale (9.0), three times, for 1{,}920 clips. The untreated clip and the scale-up clip of a pair use the same seed, so they start from the same initial noise and differ only in the scale. Each prompt and each replicate has its own seed, computed from the category, the prompt index, and the replicate and from nothing else; all 1{,}920 seeds in the released manifests follow this rule, and the two conditions share the seed in all 960 pairs. Scoring uses each suite's original judge, with rubric, frame protocol, judge model, and decoding unchanged. In ten negation pairs (three, three, and four in the three replicates) the judge marked one clip ambiguous and returned no binary label; these pairs are excluded, leaving $n=157$, $157$, and $156$, while all 160 temporal pairs are retained. The ambiguous clips fall in the graded categories (INA, NMI, DNS), where Appendix~\ref{app:kappa} already locates the judging boundary. Demand was measured separately, on the first 8 prompts of each category of all three suites (192 prompts), by reading the first quarter three times under three seeds.

\paragraph{Demand hardly depends on the seed.}
We measure this with the intraclass correlation ICC(2,1), which is 1 when the three measurements of a prompt agree perfectly and 0 when they agree no better than measurements of different prompts. Over the 192 prompts it is $.871$ (per suite: $.77$, $.76$, and $.86$ for negation, rare composition, and temporal), and any two seeds rank the prompts almost identically (Spearman correlations of $.905$ to $.920$). Reading demand from a single seed, as the rule of Fig.~\ref{fig:rule} does, is therefore justified. For temporal prompts the demand gap between failing and succeeding prompts is negative under every seed ($-.047$ to $-.089$), the direction of Table~\ref{tab:demand_sign}. For negation the gap is close to zero on this subsample ($-.006$ to $-.029$ at $n=64$ per seed) and does not show the positive sign measured on all prompts; as in Appendix~\ref{app:localization}, we rely on the statistics computed from all prompts ($n=400$, Table~\ref{tab:demand_sign}) and not on this small sample.

\paragraph{One clip is a noisy measurement of a prompt.}
Across the three seeds, a prompt receives the same label all three times for 72 to 75 percent of negation prompts and for 60 to 64 percent of temporal prompts. Fleiss' $\kappa$, a chance-corrected measure of agreement among more than two labels, is $.60$ and $.64$ for negation under the two conditions and $.47$ and $.52$ for temporal. The untreated failure rate of the same prompts moves by up to $.09$ between seeds (.431 to .525 on the temporal subset). A difference of a few points between two single runs can thus come from the seed alone.

\begin{table}[ht]
\centering
\small
\setlength{\tabcolsep}{5pt}
\begin{tabular}{llcccccc}
\toprule
Failure type & Replicate & Untreated & Scale-up & Net & Repaired & Broken & McNemar $p$ \\
\midrule
Negation & 1 & .382 & .357 & $-.025$ & 16 & 12 & .57 \\
Negation & 2 & .376 & .382 & $+.006$ & 11 & 12 & $1.0$ \\
Negation & 3 & .365 & .340 & $-.026$ & 12 & 8 & .50 \\
Negation & pooled & .374 & .360 & $-.015$ $[-.052, +.025]$ & 39 & 32 & .48 \\
\midrule
Temporal & 1 & .431 & .425 & $-.006$ & 15 & 14 & $1.0$ \\
Temporal & 2 & .525 & .450 & $-.075$ & 25 & 13 & .07 \\
Temporal & 3 & .475 & .475 & $.000$ & 19 & 19 & $1.0$ \\
Temporal & pooled & .477 & .450 & $-.027$ $[-.071, +.019]$ & 59 & 46 & .24 \\
\bottomrule
\end{tabular}
\caption{\textbf{Shared-seed replication of the two borderline scale-up cells} (Mochi, first 20 prompts per category). Repaired and broken count the pairs whose label changed between the untreated and the scale-up clip. The pooled $p$ is an exact McNemar test on the changed pairs of all three replicates together; pooling is valid because, if the scale has no effect, a changed pair is equally likely to be a repair or a break in every replicate. Brackets give 95\% bootstrap intervals on the net change, obtained by resampling prompts.}
\label{tab:seeds}
\end{table}

\paragraph{Temporal prompts: no net effect in any replicate.}
The net change is $-.006$, $-.075$, and $.000$ in the three replicates and $-.027$ pooled ($p=.24$). The bootstrap interval of $[-.071, +.019]$ means that any net effect of the raised scale on temporal prompts is small in either direction. The null of the main run is stable.

\paragraph{Negation: few repairs in every run, and no settled sign.}
The raised scale repairs little. Its recovery is $.222$ pooled here and $.171$ in the main run, against $.500$ for NEGATE, and the negation branch of Fig.~\ref{fig:rule} depends on nothing beyond this. The net change, in contrast, lies within seed noise: it is $-.025$, $+.006$, and $-.026$ in the three replicates ($-.015$ pooled, $p=.48$, interval $[-.052, +.025]$). The main run restricted to the same 160 prompts shows $+.044$, so the difference from the main run comes from the seed and not from the choice of prompts. The main run rests on 400 prompts and one seed, this replication on 160 prompts and three seeds. Together they show that the net effect of the raised scale on negation is small, with a sign that the data do not settle. What every run agrees on is what the rule uses: raising the scale does not repair negation failures.

\paragraph{The results do not depend on where the judge draws its line.}
The judge here is the same as in the main runs (judge model, rubric, frame protocol, and decoding unchanged), so the two sets of runs are measured on one scale. The negation judge returns a graded score and a verdict for every clip, which lets us relabel the clips in four other ways: by the verdict, by a strict cut that counts scores $\le 2$ as failures, by a lenient cut at $\le 3$ that also counts ambiguous clips as failures, and by the judge's yes-or-no answer on whether forbidden content is present. Recomputing Table~\ref{tab:seeds} under each of the four moves the pooled negation net change only between $-.015$ and $-.019$. The temporal cell stays near zero under both of its labelings (the judge's failure flag and a score cut). Clips the judge marked ambiguous are 0.8 percent of the total, and including or excluding them changes no number above at the reported precision.

\paragraph{The rare-composition cell replicates as well.}
The dose-response runs of Appendix~\ref{app:dose} later added shared-seed replicates for rare composition, which is not a borderline cell. Pooled over the three replicates, the direction of Table~\ref{tab:grid} holds, with 60 pairs repaired against 31 broken (exact $p=3.1\times10^{-3}$); the per-replicate counts are 19/10, 28/6, and 13/15, the third on its own within noise.

\section{The Recovery Curves under a Re-seed Control}
\label{app:reseed}

Recovery is computed over the prompts that failed untreated, and in the main protocol the treated clip of a prompt uses a different seed from the untreated one. Some of those prompts would therefore succeed on a second draw even if the method did nothing. This re-sampling luck is also not spread evenly along demand. In rare composition, prompts with high demand fail less often, so the failures found at high demand include more prompts that failed by accident. The rise of recovery with demand in Fig.~\ref{fig:boundary} could in principle come from this effect alone. Two measurements show that it does not.

\begin{figure}[ht]
    \centering
    \includegraphics[width=0.9\linewidth]{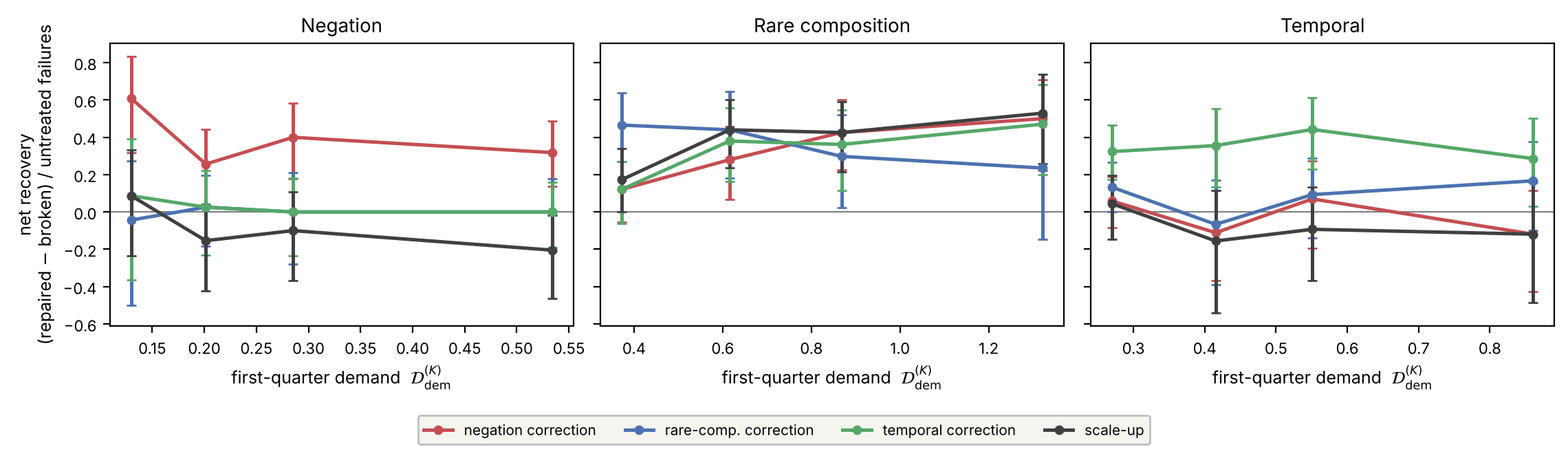}
    \caption{\textbf{Net recovery by demand quartile, computed over all prompts.} Net recovery is the number of repaired prompts minus the number of broken prompts, divided by the number of untreated failures. Re-sampling alone would keep it at zero in every bin (zero line), so a curve that lies above zero and rises with demand shows an effect of the method and not of selection. In the legend, the negation, rare-composition, and temporal corrections are NEGATE, DCR, and TPD. Error bars are 95\% bootstrap intervals, obtained by resampling prompts.}
    \label{fig:netrecovery}
\end{figure}

\paragraph{Net recovery is zero under re-sampling and rises with demand under the methods.}
For each demand bin we take all 400 prompts of a suite, not only the failures, and count the prompts that a method repaired and the prompts that it broke. Net recovery is the difference of the two counts divided by the number of untreated failures. If a method did nothing beyond re-sampling, a prompt would be as likely to change from failure to success as from success to failure, so net recovery would be zero on average in every bin, however the failure rate varies with demand. The measured values rise with demand (Fig.~\ref{fig:netrecovery}). For the raised scale on rare composition, net recovery goes from $+.17$ in the lowest demand quartile to $+.53$ in the highest, for NEGATE from $+.12$ to $+.50$, and for TPD from $+.12$ to $+.47$. The rise from the lowest to the highest quartile is $+.36$ for the raised scale (bootstrap 95\% interval $[+.06,\, +.64]$), $+.38$ for NEGATE ($[+.07,\, +.66]$), and $+.35$ for TPD ($[+.03,\, +.62]$), from 5{,}000 resamples of prompts with the quartile edges held fixed. In the lowest quartile the per-bin intervals of all three methods include zero ($[-.02,\, +.33]$ for the raised scale), matching the main text's reading of that bin, and from the second quartile on every interval excludes it. Other binnings give the same picture (for the raised scale, $+.21/+.49/+.46$ by tercile and $+.30/+.47$ by median split). Neither the rise nor the level beyond the lowest quartile can therefore be due to selection. DCR, the correction designed for this type, behaves differently: its net recovery is above zero in every quartile ($+.47/+.44/+.30/+.24$ from low to high demand) and does not rise with demand (quartile difference $-.23$, bootstrap interval $[-.65,\, +.12]$), so the dependence on demand belongs to the raised scale and to the corrections not designed for the type.
 For negation, the net recovery of the raised scale is negative in the three upper demand quartiles ($+.09/-.15/-.10/-.21$), which is where the reaction to the negated content is strongest. These bins re-cut the single main run whose negation net change varies in sign with the seed (Appendix~\ref{app:seeds}), so we read them as consistent with the mechanism and not as independent evidence of harm.

\paragraph{The ineffective cells recover at the re-seed rate, and the effective corrections exceed it.}
The shared-seed runs of Appendix~\ref{app:seeds} contain three untreated replicates of the same prompts for negation and for temporal prompts, so they show directly how much re-sampling alone repairs. Among the prompts that fail in one untreated replicate, we take the fraction that succeed in another, over all six ordered pairs of replicates. This re-seed rate is $.250$ for negation and $.279$ for temporal prompts. It is measured on the 160-prompt subsets of Appendix~\ref{app:seeds}, whereas the cells of Table~\ref{tab:grid} use all 400 prompts. Its dependence on demand follows the directions of Table~\ref{tab:demand_sign}: it falls with demand for negation ($.31/.25/.21$ by tercile), where high-demand failures are persistent, and rises for temporal prompts ($.15/.35/.42$), where high-demand failures are more often accidental. The untreated replicates generated for Appendix~\ref{app:dose} extend this control to rare composition, where the re-seed rate is $.325$ and is lowest at low demand ($.24/.40/.38$ by tercile).

\begin{figure}[ht]
    \centering
    \includegraphics[width=0.92\linewidth]{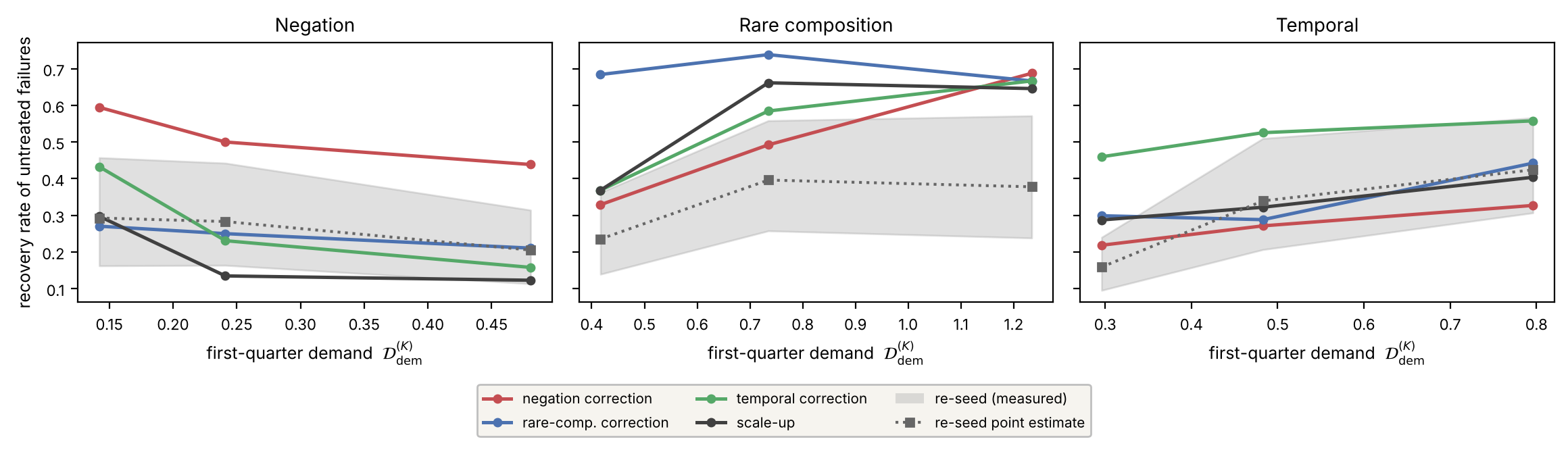}
    \caption{\textbf{Recovery against demand with a measured re-seed control.} The gray band is the recovery that re-sampling alone achieves, measured from three untreated replicates (negation and temporal from Appendix~\ref{app:seeds}, rare composition from Appendix~\ref{app:dose}; 95\% bootstrap interval, resampling prompts), and the dotted line is its point estimate. Only the cells that Table~\ref{tab:grid} finds effective lie above the band; on rare composition every method rises above it at high demand. In the legend, the negation, rare-composition, and temporal corrections are NEGATE, DCR, and TPD.}
    \label{fig:reseedcontrol}
\end{figure}

Figure~\ref{fig:reseedcontrol} compares the cells of Table~\ref{tab:grid} with this control. The cells that the table finds ineffective recover at about the re-seed rate: DCR at $.240$ and TPD at $.253$ on negation against $.250$, and NEGATE at $.263$, the raised scale at $.328$, and DCR at $.333$ on temporal prompts against $.279$. What the grid calls no effect is thus the recovery that re-sampling gives. The effective corrections exceed it. NEGATE recovers twice the re-seed rate ($.500$) and lies above the control band in every demand tercile, and TPD ($.505$) lies above the band over most of the demand range. The raised scale on negation recovers $.171$, which does not exceed the re-seed rate of $.250$: on negations it repairs nothing beyond what generating again would, in agreement with Appendix~\ref{app:seeds}. On rare composition the picture is the reverse: every method sits at the band in the lowest demand bin and rises above it from the breakpoint on (middle panels of Figs.~\ref{fig:boundary} and~\ref{fig:reseedcontrol}).

\paragraph{Recovery against demand, per failure type.}
This paragraph gives the analysis behind Sec.~\ref{subsec:exp_boundary}. In rare composition, the recovery of the three methods not designed for the type rises with demand ($r = +0.22$ to $+0.31$, $p \le 0.002$). The segmented fit places the breakpoint at $D^{*}=0.59$ with a wide interval ($[0.49, 1.14]$; Table~\ref{tab:segmented}), and each method's separately fitted threshold lies inside it, so we rely on the direction of the trend and not on the exact threshold value. For negation the trend is reversed and recovery falls as demand rises ($r = -0.09$ to $-0.18$), while for temporal prompts it rises slightly ($r = +0.18$, $p=.013$). These two trends are not correction effects. Re-sampling alone has the same two slopes, and the methods not designed for these two types recover at about the re-seed rate over the whole demand range. The trends describe the failures themselves, which are persistent at high demand for negation and more often accidental at high demand for temporal prompts, so the breakpoints fitted for these two types (negation .212, temporal .316) are descriptive only.

\paragraph{Choosing between DCR and the raised scale.}
The decision rule has to choose between two methods on rare composition, and this is the only place where we compare one method with another. Both are paired with the same untreated clips, and we compare them on net recovery, which also counts the prompts a method breaks. Over the whole suite the two lower the failure rate equally (by .178 and .175), so neither is better overall; they differ by demand. Below the threshold the raised scale and the two corrections designed for other types repair only .33 to .39 of failures (the raised scale .385), close to the re-seed rate. Above the threshold DCR repairs slightly more failures than the raised scale (.703 against .649) but breaks more prompts that had succeeded (induction .276 against .132), so the raised scale has the larger net recovery ($+.47$ against $+.32$). Below the threshold DCR does ($+.45$ against $+.23$). The difference between the two methods changes sign at the threshold (contrast $+.36$ between the above- and below-threshold differences, bootstrap 95\% interval $[+.09,\, +.64]$), so reading demand changes the choice.

\paragraph{The demand bins are stable across seeds.}
The analyses above place prompts in demand bins, so the bins must not depend on the seed. On the 192-prompt sample with three demand reads under different seeds, a prompt falls in the same tercile as under the main-run read 64 to 88 percent of the time, and the rank correlation with the main-run read is $.79$ to $.93$. Together with the ICC of $.871$ (Appendix~\ref{app:seeds}), this shows that the demand axis of Fig.~\ref{fig:boundary} reflects the prompts and not the seed.

\section{Dose-Response of the Guidance Scale}
\label{app:dose}

The main protocol compares two values of the guidance scale, the default 4.5 and the published raise to 9.0. This appendix measures the response along a geometric grid of five scales, $3.2 / 4.5 / 6.4 / 9.0 / 12.7$, spaced by a factor $\sqrt{2}$ so that the published pair lies exactly on the grid, on the fixed 160-prompt subsets of Appendix~\ref{app:seeds}. Every scale of a prompt starts from the initial latent of its untreated run under the seed rule of that appendix, so each dose pairs exactly, prompt by prompt, with the same-seed clip at 4.5, and the counts below are within-pair changes with no seed noise between conditions. Rare composition had no shared-seed runs, so it received everything: three replicates at 4.5 and at 9.0, and two replicates at each of 3.2, 6.4, and 12.7. Negation and temporal prompts already have three replicates at 4.5 and 9.0 (Appendix~\ref{app:seeds}), so they received one replicate at each of the two new raised scales, 6.4 and 12.7; they were not run at 3.2.
 Scoring uses each suite's original judge, unchanged, over 2{,}560 new clips. The subset's untreated rates differ from the 400-prompt main run (rare composition .346 against .472), so this appendix compares only within its own pairs, and its rates are not to be read against the main tables.

\begin{table}[ht]
\centering
\small
\setlength{\tabcolsep}{5pt}
\begin{tabular}{llccccc}
\toprule
Failure type & Scale $w$ & Pairs & Recovery & Induction & Net recovery [95\% CI] & McNemar $p$ \\
\midrule
Rare comp. & 3.2  & 320 & .231 & .138 & $-.009$ $[-.153,\, +.115]$ & $1.0$ \\
Rare comp. & 6.4  & 320 & .333 & .069 & $+.214$ $[+.099,\, +.325]$ & $8.0\times10^{-4}$ \\
Rare comp. & 9.0  & 480 & .361 & .099 & $+.175$ $[+.061,\, +.288]$ & $3.1\times10^{-3}$ \\
Rare comp. & 12.7 & 320 & .419 & .059 & $+.316$ $[+.187,\, +.438]$ & $2.0\times10^{-6}$ \\
\midrule
Negation & 6.4  & 156 & .133 & .083 & $.000$ $[-.140,\, +.122]$ & $1.0$ \\
Negation & 9.0  & 470 & .222 & .109 & $+.040$ $[-.067,\, +.142]$ & $.48$ \\
Negation & 12.7 & 156 & .267 & .156 & $+.017$ $[-.182,\, +.182]$ & $1.0$ \\
\midrule
Temporal & 6.4  & 160 & .145 & .121 & $-.014$ $[-.155,\, +.111]$ & $1.0$ \\
Temporal & 9.0  & 480 & .258 & .183 & $+.057$ $[-.040,\, +.146]$ & $.24$ \\
Temporal & 12.7 & 160 & .261 & .176 & $+.029$ $[-.149,\, +.182]$ & $.86$ \\
\bottomrule
\end{tabular}
\caption{\textbf{The guidance scale along a geometric grid}, against the same-seed clip at the default scale 4.5. Net recovery as in Appendix~\ref{app:reseed} (repaired minus broken pairs, over untreated failures; positive is an improvement), with prompt-cluster bootstrap intervals (5{,}000 resamples). The 9.0 rows pool the three shared-seed replicates (for negation and temporal prompts, those of Appendix~\ref{app:seeds}).}
\label{tab:dose}
\end{table}

\paragraph{Rare composition benefits from every raise, and the data do not separate 9.0 from 12.7.}
Net recovery is $-.009$ at 3.2, $+.214$ at 6.4, $+.175$ at 9.0, and $+.316$ at 12.7 (Table~\ref{tab:dose}). It is significant at every raised scale and absent at the lowered one. Recovery itself rises with each step (.231, .333, .361, .419). The step from 9.0 to 12.7 repairs 22 pairs and breaks 16 ($p=.42$), so the data do not distinguish the two scales. The point estimate is highest at 12.7, and a further gain beyond the published raise is neither shown nor excluded at this sample size. Lowering the scale to 3.2 has no net effect overall, and above the demand threshold its point estimate is negative ($-.090$, against $+.100$ below; neither significant at these sizes), the direction the mechanism predicts, since high-demand prompts are the ones whose weak reaction guidance can still amplify. No raised scale in the tested range reverses the benefit; if Mochi has an over-amplification regime of the kind CogVideoX enters already at a factor 1.7 (Appendix~\ref{app:cogscale}), it lies beyond a factor 2.8.

\paragraph{No scale helps negation.}
The net recovery of the raised scale on negation is $.000$ at 6.4, $+.040$ at 9.0, and $+.017$ at 12.7, all within seed noise ($p \ge .48$). The negation branch of Fig.~\ref{fig:rule} therefore does not depend on the choice of 9.0. A gentler raise does not help either, since at 6.4 the scale repairs exactly as many pairs as it breaks (8 against 8). The flat curve also limits what can be said about the mechanism. If a larger scale added more of the forbidden content, failures would grow with the scale, and they do not. What the data support is the weaker statement that the scale, at any strength, does not correct a reaction aimed at the forbidden content.

\paragraph{No scale has a net effect on temporal prompts.}
Net recovery is $-.014$, $+.057$, and $+.029$ at the three raised scales ($p \ge .24$). A scalar applied equally to every frame does not place an event in time at any strength.

\paragraph{Two by-products: a re-seed control for rare composition and a replication of its 9.0 cell.}
The three untreated replicates show how much re-sampling alone repairs in rare composition. This re-seed rate is $.325$ ($.24/.40/.38$ by demand tercile, lowest at low demand), and it is the gray band of the rare-composition panels in Figs.~\ref{fig:boundary} and~\ref{fig:reseedcontrol}. The 9.0 runs repeat the main-run cell with shared seeds, and its direction holds in the pooled count, with 60 pairs repaired and 31 broken over the three replicates (exact $p=3.1\times10^{-3}$; per-replicate counts in Appendix~\ref{app:seeds}).

\paragraph{Colors saturate as the scale rises, and generation does not break down.}
The mean HSV saturation of the rare-composition clips rises with every step of the grid ($.346/.400/.437/.467/.474$) and flattens after 9.0, and the Hasler--Suesstrunk colorfulness rises from 45 to 75. This is the side effect visible in Fig.~\ref{fig:rx_rarecomp}, measured along the scale. Near-uniform breakdowns remain rare at every scale (at 12.7, one clip of 320), so Mochi tolerates a factor 2.8 where CogVideoX degenerates at 1.7, and the inverted rare-composition response of Appendix~\ref{app:cogscale} is not explained by a shared breakdown point.

\begin{figure}[ht]
    \centering
    \includegraphics[width=0.95\linewidth]{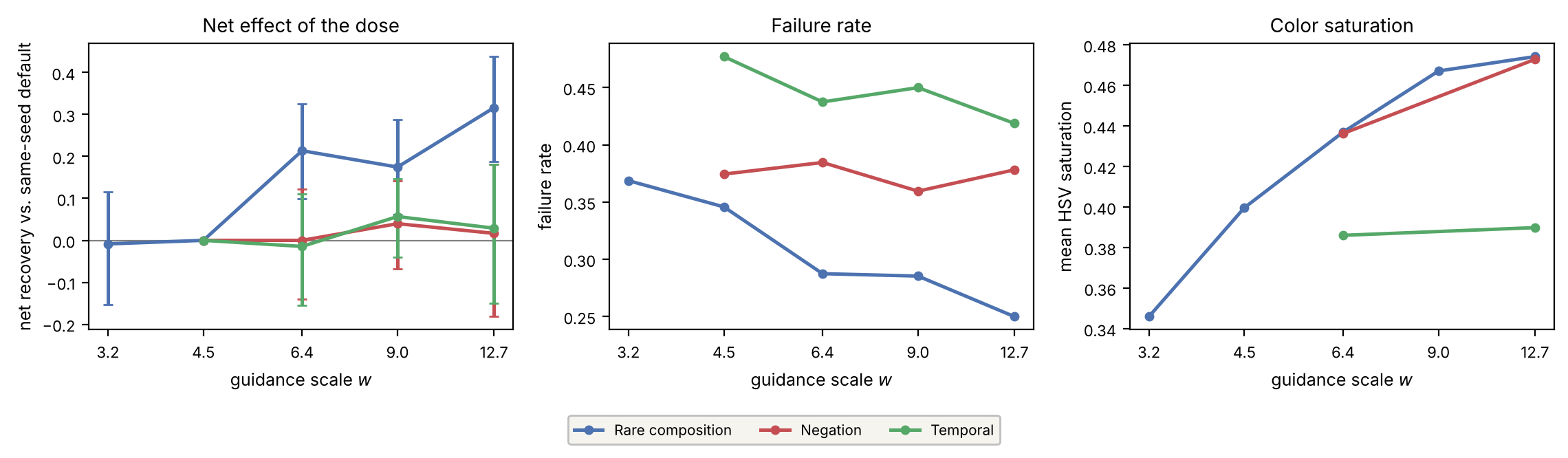}
    \caption{\textbf{Dose-response of the guidance scale} on the 160-prompt subsets. Left: net recovery against the same-seed clip at the default scale; the zero line is the expectation if the scale did nothing, and error bars are prompt-cluster bootstrap 95\% intervals. Middle: failure rate, pooled over replicates. Right: mean HSV saturation of the clips (for negation and temporal prompts, measured at the two new scales). Only rare composition responds, at every raised scale, and its color saturation grows with the scale.}
    \label{fig:dose}
\end{figure}

\section{Prompts That No Method Repairs}
\label{app:switch}\label{subsec:exp_switch}

\begin{figure}[ht]
    \centering
    \includegraphics[width=0.6\linewidth]{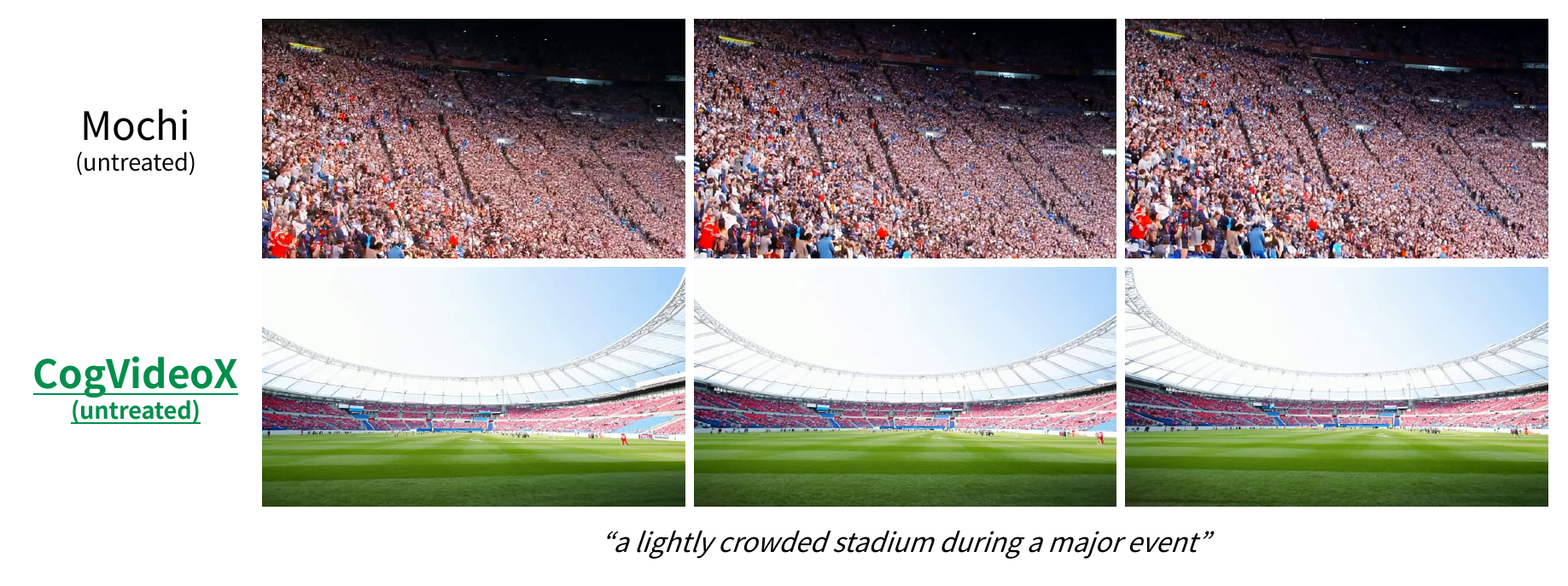}
    \caption{\textbf{A prompt that no correction reaches on one model succeeds untreated on another.} On Mochi this prompt's demand is 0.21, below its failure type's threshold, and all four methods fail. CogVideoX registers the same request at 0.55, and its untreated clip shows the lightly crowded stadium (bottom, \rx{green underlined}). Seventeen prompts in this failure type resist all four methods; twelve succeed untreated on another model. That rate matches what the other models' overall success rates already imply, so we report it as an observation about model-specificity and not as a selection rule.}
    \label{fig:switch}
\end{figure}

Seventeen rare-composition prompts, 9\% of the untreated rare-composition failures on Mochi, resist all four methods. Twelve of the seventeen succeed, untreated, on at least one of the other two models (Fig.~\ref{fig:switch}). A prompt can therefore be out of reach of every guidance-based correction on one model and need no correction on another. We do not turn this into a method for choosing a model, for two reasons. First, the two other models succeed on roughly 81\% of arbitrary prompts in this failure type (from their untreated rates .375 and .518, assuming independence), so 12 of 17 is no better than expected. Second, choosing between the two by measured demand succeeds on 8 of the 17 prompts, while always choosing the stronger model succeeds on 10. At this sample size demand does not select the model. Some failures, then, are limits of one model and not of the prompt: when no correction works, trying another model is reasonable, but demand cannot yet say which one.

\section{Why the Relation Between Demand and Recovery Is Not Circular}
\label{app:identification}

Prior methods construct directions from the same denoiser, but to correct, and each differently: a forbidden span in~\cite{kang2026negate}, a relaxed counterpart in~\cite{kang2026dcr}, a truncated segment in~\cite{kang2026tpd}. These constructions are not interchangeable and we do not claim they are. The default prompt (Eq.~\ref{eq:default_operator}) is a measurement construct: one reference point, applied identically across failure types, against which outputs and predictions become comparable. Its adequacy is tested and not assumed, in two ways: generating from $\pd$ must reproduce the failure observed under $p$ (Sec.~\ref{subsec:default_operator}), and the unified judgment must agree with each suite's original judge (Appendix~\ref{app:kappa}).

The possible circularity concerns temporal prompts and TPD. For a temporal prompt, the default prompt (the request with its late event removed) is the same text that TPD itself uses as its reference. TPD's correction pushes the generation along the difference between the predictions for the two prompts, and demand is the size of that same difference, $d_t$. The overlap is intended, since the correction acts on exactly the quantity that we measure. It does, however, raise an objection. If demand is the size of a vector and TPD works by adding that vector, then finding that demand predicts TPD's success might be a consequence of the arithmetic and tell us nothing about the model.

Three facts show that it is not. First, demand and recovery come from different runs. Demand is computed on untreated trajectories, with no correction applied. Recovery is determined afterwards from the finished, corrected video and a judge's label. Nothing in the arithmetic links the two, so a relation between them has to be found in the data. Second, the relation that we interpret appears where the overlap does not exist. Our claim concerns rare composition, where NEGATE and TPD add a direction that differs from $d_t$ by construction and the raised scale adds no direction at all; the net recovery of all three rises with demand there (Fig.~\ref{fig:netrecovery}). We do not use the four cells of corrections applied to negation and temporal prompts as evidence, because their trends are those of re-sampling (Appendix~\ref{app:reseed}). In the overlapping cell itself there is little to explain: the net recovery of TPD on temporal prompts is roughly level across the demand quartiles (Fig.~\ref{fig:netrecovery}). Third, the transfer analysis (Eq.~\ref{eq:transfer}) shows the same demand predicting untreated failure on models on which no correction is ever run. The arithmetic objection can explain neither the second nor the third fact. All three are explained if demand reflects how strongly the model reacts to the difficult part and if that reaction determines how much a correction can achieve.

\section{Where Demand Sits in Denoising Time and in Clip Time}
\label{app:localization}

The profiles in this section come from a sample of 47 prompts for which we recorded the full trajectory.

\subsection{Denoising time: failure is settled early}

\begin{figure}[ht]
    \centering
    \includegraphics[width=\linewidth]{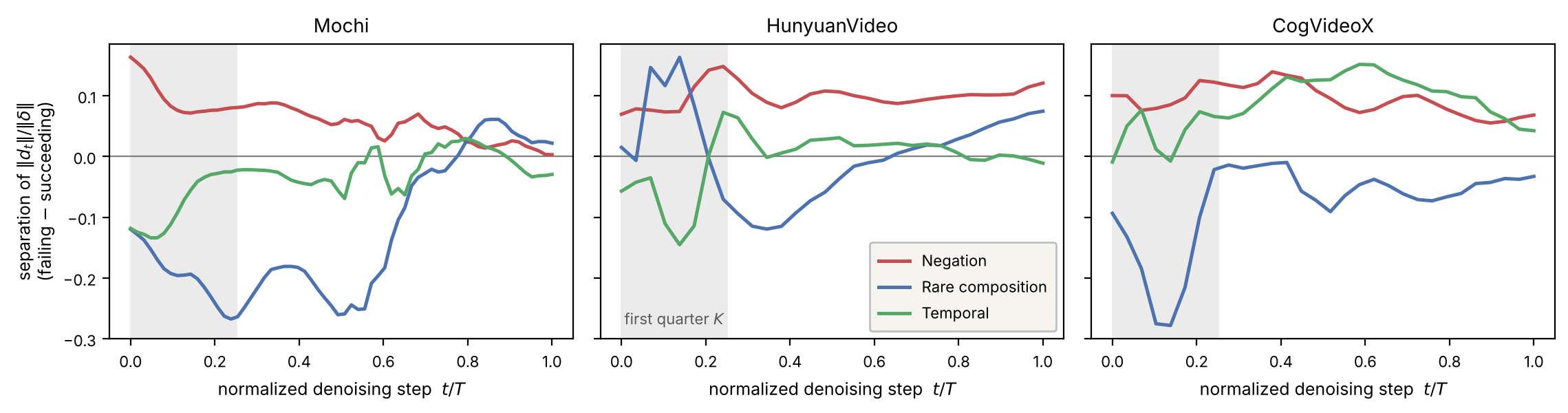}
    \caption{\textbf{Where in denoising failing and succeeding prompts part.} Difference in the per-step demand ratio $\|d_t\|/\|\epsilon_{\text{text}}-\epsilon_{\text{uncond}}\|$ between failing and succeeding prompts (positive means failing prompts react more), per failure type, on the 47-prompt sample. The shaded band is the first quarter of the steps. Negation is above zero on every model and rare composition below, and the difference is established during structure formation.}
    \label{fig:profiles}
\end{figure}

Figure~\ref{fig:profiles} plots, at each denoising step, the difference in the demand ratio between failing and succeeding prompts.

The directions of Table~\ref{tab:demand_sign} are not an artifact of averaging over the early steps. For negation the difference is above zero for essentially the whole trajectory on every model. For rare composition it is below zero through the structure-forming phase.

The difference also appears early in denoising. On Mochi, the curve for rare composition reaches its lowest point within the shaded first quarter of the steps; that is, the gap between failing and succeeding prompts is already as large there as it ever becomes. (The curve has one low point near one quarter of the steps and a second of similar depth near one half.) The gap then shrinks, closes at about three quarters of the trajectory, and over the final quarter failing and succeeding prompts cannot be told apart by demand. On CogVideoX the lowest point likewise falls within the first quarter. The early steps of denoising are those in which the overall layout of the video is formed, and the later steps add detail, so the curves indicate that whether a prompt will fail is settled while the layout is being formed. This also explains why demand computed from the first quarter alone works: it is computed in the phase in which failing and succeeding prompts differ most, and it ranks prompts almost exactly as demand over all steps does (Table~\ref{tab:prefix}).

Two exceptions on this small sample should be noted. On HunyuanVideo the rare-composition gap opens at the boundary of the first quarter and not strictly inside it. On CogVideoX the curve for temporal prompts lies above zero over most of the trajectory, the opposite of what the full data show (a correlation of $-0.043$ over all prompts). In both cases we rely on the statistics computed from all prompts (Table~\ref{tab:demand_sign}, $n{=}400$ per failure type) and not on this sample, which contains only sixteen temporal prompts and is too small to settle the question.

\subsection{Clip time: demand is largest where the prompt places the event}

\begin{figure}[ht]
    \centering
    \includegraphics[width=\linewidth]{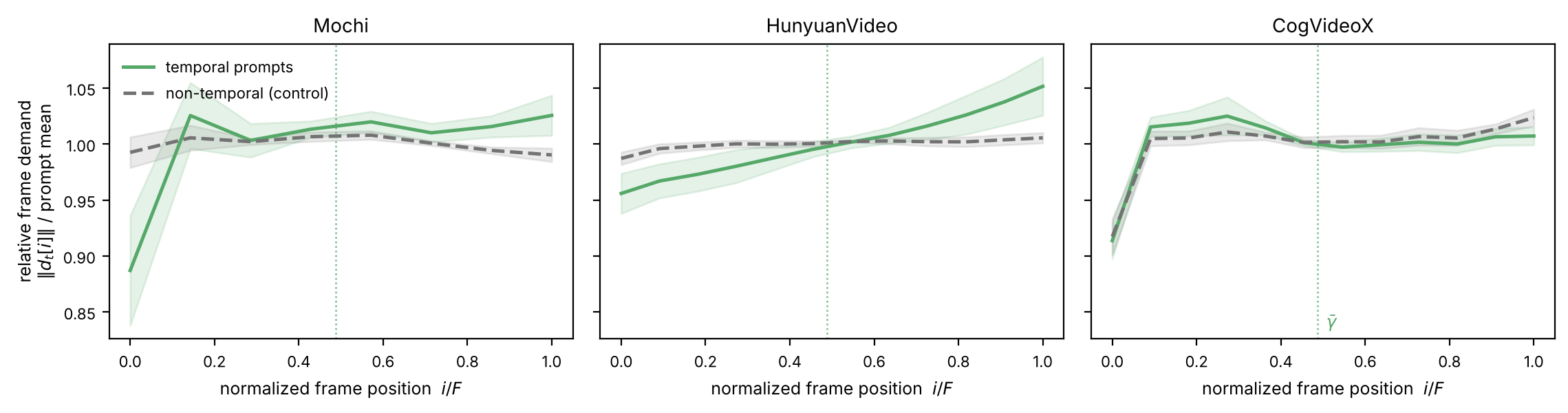}
    \caption{\textbf{Where in the clip the model is asked to change something.} Relative per-frame demand $\|d_t[i]\|$ (first-quarter steps, normalized per prompt) at normalized frame position $i/F$, for temporal prompts and for the pooled non-temporal control. The dotted line marks the mean event boundary $\bar{\gamma}=0.49$.}
    \label{fig:frames}
\end{figure}

Figure~\ref{fig:frames} resolves first-quarter demand by frame. For temporal prompts, the per-frame demand rises toward the late part of the clip on Mochi and on HunyuanVideo. On HunyuanVideo the profile crosses the non-temporal control near the mean event boundary $\bar{\gamma}=0.49$; on Mochi it crosses early, near $i/F \approx 0.1$, and stays slightly above the control, rising further over the last frames. On CogVideoX the temporal profile shows no late rise and is indistinguishable from the control.

The frame axis locates the \emph{request}, not the failure. Whether per-frame profiles also separate failing from succeeding clips depends on the model at this sample size, and we do not claim it.

\subsection{Alignment}

Alongside demand we record the alignment
\begin{equation}
\cos\theta_t \;=\; \frac{d_t^{\top}\,\delta_{\text{ref}}}{\|d_t\|_2\,\|\delta_{\text{ref}}\|_2},
\label{eq:alignment}
\end{equation}
the cosine between the direction $d_t$ of the model's reaction to the difficult part and the applied guidance update $\delta_{\text{ref}}$. In the main analyses it enters only as a covariate in the per-type model of Table~\ref{tab:demand}.

\section{The Five Checks in Detail}
\label{app:checks}

This section gives the full version of the checks summarized in Sec.~\ref{sec:checks}. The models and statistics behind them are specified in Appendices~\ref{app:formal} and~\ref{app:tables}.

\paragraph{Text-only baselines.}\label{subsec:exp_atlas}\label{subsec:exp_complexity}
A first concern is that the prompt text alone might predict failure. Failure is frequent on every model (rates of .365 to .688 per failure type) and varies widely: of the 72 type-category-model cells, five are at or below .10 and four at or above .90 (Appendix~\ref{app:atlas}). To test this, we tried to predict which prompts fail using only quantities that can be computed from the prompt text, without running the video model. We used three kinds of quantity: five measures of how complex the prompt is (its length, the numbers of words and of entities, the number of modifiers, and the depth of its sentence structure); a rating $r$, given by a language model, of how rare the described scene is in the real world; and the distance $u$ between $p$ and $\pd$ in the text encoder that the video model uses (definitions in Appendix~\ref{app:formal}). We score each predictor by its AUROC, which answers the following question: if we pick one failing prompt and one succeeding prompt at random, how often does the predictor give the failing one the higher score? A value of .500 means the predictor is no better than guessing. With all text-only predictors combined, AUROC is 0.607/0.635/0.694 on the three models when the three failure types are pooled. These pooled values overstate what the predictors can do, because the pooled model also knows which failure type each prompt belongs to, and the failure types differ in how often they fail. Within a single failure type the predictors do much worse: for rare composition they are at the level of guessing on two of three models (.520, .523), and among temporal prompts they stay below .65 throughout. The rarity rating fails because the suite holds real-world rarity nearly constant (63\% of rare-composition prompts are rated 6 or 7 of 7) while failure varies from prompt to prompt. The encoder distance $u$ is the closest text-only counterpart of demand: it compares the same two prompts, $p$ and $\pd$, in the same text encoder, and differs from demand only in never running the video denoiser. Within rare composition and within temporal prompts its AUROC is .44 to .55, no better than guessing. What demand captures is therefore not present in the text encoder's representation of the prompt; it arises only when the denoiser processes the prompt. In negation the text-only predictors appear to work well (.771 to .824). The reason is that the negation suite has eight categories with very different failure rates, and the wording of a prompt reveals its category, so the predictors are recognizing the category and not separating failing from succeeding prompts within one. \emph{In short, the text of a prompt does not tell us whether it will fail; that has to be read from the model.}

\paragraph{Added value of demand.}
Demand adds predictive power beyond these baselines, though modestly (Table~\ref{tab:demand_main}b). Added to all text-only predictors with a single slope, demand gains $+0.025$, $+0.056$, $+0.001$ AUROC on the three models. A single slope averages a positive correlation (negation) with negative ones (the other two), so it understates the signal, and on CogVideoX the two cancel. With one slope per failure type the gain is $+0.050$, $+0.104$, $+0.011$. The cleanest test is inside rare composition with category fixed effects, where text alone reaches .588/.748/.689 and adding demand gives .628/.782/.739, a gain on all three models. Excluding clips the judge found ambiguous reproduces the pattern ($+.026/+.056/+.002$ pooled). We do not propose demand as a standalone classifier: whether a given prompt fails also depends on the random seed, which caps what any prompt-level score can reach. \emph{Demand thus carries information that no text-only predictor has. It is best used, together with the failure type, to decide what to try after a failure, and not as a detector that says in advance whether a single prompt will fail.}

\paragraph{Transfer across models.}\label{subsec:exp_transfer}
Demand measured on one model predicts untreated failure on another, even though which prompts fail is partly model-specific: the correlation $\phi$ of per-prompt failure between two models is .48 for negation and only .16 to .26 for rare composition, so whether \emph{a snowy beach} fails depends on which model is asked. In the pooled transfer matrix (Table~\ref{tab:transfer}), every off-diagonal AUROC lies within 0.03 of the same-model value in its column (0.577/0.617/0.631). \emph{Demand therefore measures something that the models share, whereas the conditions under which raising the scale helps differ from model to model (Appendix~\ref{app:cogscale}). For this reason both should be measured on the model that is actually in use.}

\paragraph{Circularity and judge agreement.}
One could worry that the link between demand and recovery is circular for TPD, whose correction direction coincides with $\epsilon_{\text{text}} - \epsilon^{\circ}$ for temporal prompts. Three facts rule it out: demand is measured on untreated runs and recovery on treated outputs; the same relation holds for corrections whose direction differs from that difference and for scale-up, which adds no direction at all; and demand predicts untreated failure on models where no correction ever runs (Appendix~\ref{app:identification}). The relation is an empirical result and not a by-product of how one correction is built.

\label{subsec:exp_validity} A last concern is that the three suites use three judges. To check that they measure one thing, a single judge asked one question of every clip: does it show $\pd$ rather than $p$? We compared the two sets of labels on 201 clips, chosen so that every model, failure type, category, and label is represented (a stratified sample). Agreement is measured with Cohen's $\kappa$, which is 0 when two judges agree no more often than they would by chance and 1 when they always agree. We obtain $\kappa = 0.49$ overall (0.41 to 0.59 per failure type), a moderate level of agreement. To understand the disagreements, we use the fact that the original judges also give each clip a graded score before deciding between failure and success. Clips on which the two judgments disagree have a middle score far more often than clips on which they agree (49\% of disagreeing clips against 23\% of agreeing ones; Mann-Whitney test, which compares two groups of scores without assuming how they are distributed, $p=1.1\times10^{-4}$). In other words, the judgments disagree mainly on borderline clips, in which the video partly follows the prompt and a yes-or-no label is hard to assign. The two judgments thus measure the same thing and differ only in where they draw the line on such clips, so all analyses above use each suite's original judgment (Appendix~\ref{app:kappa}). The three suites' labels therefore answer the same question and failure rates can be compared across failure types, although borderline clips remain a source of noise.

\section{Detailed Results}
\label{app:tables}

\subsection{Text-only predictors}

Table~\ref{tab:complexity} gives the AUROC of every text-only predictor, alone and combined. These values are the baseline that demand has to improve on.

\begin{table}[ht]
\centering
\small
\begin{tabular}{lcccc@{\hspace{1.6em}}ccc}
\toprule
& \multicolumn{4}{c}{Pooled, failure-type fixed effects} & \multicolumn{3}{c}{Within one failure type, all predictors} \\
\cmidrule(r){2-5}\cmidrule(l){6-8}
Model & Complexity & Rarity $r$ & Encoder $u$ & All & Negation & Rare comp. & Temporal \\
\midrule
Mochi        & 0.569 & 0.595 & 0.545 & 0.607 & 0.819 & 0.523 & 0.637 \\
HunyuanVideo & 0.623 & 0.617 & 0.623 & 0.635 & 0.824 & 0.705 & 0.583 \\
CogVideoX    & 0.660 & 0.669 & 0.664 & 0.694 & 0.771 & 0.520 & 0.648 \\
\bottomrule
\end{tabular}
\caption{\textbf{Predicting failure from the prompt text alone.} Five-fold cross-validated AUROC of Eq.~\eqref{eq:regression} without the demand term (chance $=0.500$): five complexity measures, the rarity rating $r$, and the encoder distance $u$, alone and combined. Left: pooled over failure types with failure-type fixed effects. Right: refit within each failure type, where fixed effects cannot help.}
\label{tab:complexity}
\end{table}

\subsection{What demand adds}

Table~\ref{tab:demand} gives the full version of Table~\ref{tab:demand_main}b, and Table~\ref{tab:demand_sign} repeats the directions of Table~\ref{tab:demand_main}a. Excluding the clips the judge found ambiguous leaves the pooled gain unchanged: $+.026/+.056/+.002$ on Mochi/HunyuanVideo/CogVideoX.

\begin{table}[ht]
\centering
\small
\setlength{\tabcolsep}{5pt}
\begin{tabular}{lccc@{\hspace{1.2em}}cc@{\hspace{1.2em}}ccc}
\toprule
& \multicolumn{3}{c}{Pooled, one demand slope} & \multicolumn{2}{c}{One slope per failure type} & \multicolumn{3}{c}{Rare comp., category fixed effects} \\
\cmidrule(r){2-4}\cmidrule(r){5-6}\cmidrule(l){7-9}
Model & Text only & $+$ demand & $\Delta$ & AUROC & $\Delta$ & Text only & $+$ demand & $\Delta$ \\
\midrule
Mochi        & 0.607 & 0.632 & $+0.025$ & 0.658 & $+0.050$ & 0.588 & 0.628 & $+0.040$ \\
HunyuanVideo & 0.635 & 0.691 & $+0.056$ & 0.739 & $+0.104$ & 0.748 & 0.782 & $+0.035$ \\
CogVideoX    & 0.694 & 0.695 & $+0.001$ & 0.706 & $+0.011$ & 0.689 & 0.739 & $+0.049$ \\
\bottomrule
\end{tabular}
\caption{\textbf{AUROC gained when demand is added} (five-fold cross-validation). Left: all text-only predictors with failure-type fixed effects (Eq.~\ref{eq:regression}), against the same model plus one linear demand term. Center: demand and alignment slopes fitted per failure type, the form that the opposite directions of Table~\ref{tab:demand_sign} call for. Right: within rare composition with category fixed effects; without those fixed effects the text-only predictors are at chance there on two of three models (Table~\ref{tab:complexity}).}
\label{tab:demand}
\end{table}

\begin{table}[ht]
\begin{minipage}[t]{0.47\linewidth}
\centering
\small
\begin{tabular}{lccc}
\toprule
& Negation & Rare comp. & Temporal \\
\midrule
Mochi        & $+0.092$ & $-0.176$ & $-0.116$ \\
HunyuanVideo & $+0.290$ & $-0.345$ & $-0.267$ \\
CogVideoX    & $+0.148$ & $-0.070$ & $-0.043$ \\
\bottomrule
\end{tabular}
\caption{\textbf{Direction of the relation between demand and failure.} Point-biserial correlation between the failure label and first-quarter demand, per model and failure type. Demand itself is nonnegative (Eq.~\ref{eq:demand}); the sign belongs to the correlation.}
\label{tab:demand_sign}
\end{minipage}\hfill
\begin{minipage}[t]{0.47\linewidth}
\centering
\small
\begin{tabular}{lccc}
\toprule
& \multicolumn{3}{c}{Failure predicted on} \\
\cmidrule(l){2-4}
Demand measured on & Mochi & Hunyuan & CogVX \\
\midrule
Mochi        & 0.577 & 0.587 & 0.630 \\
HunyuanVideo & 0.580 & 0.617 & 0.636 \\
CogVideoX    & 0.571 & 0.592 & 0.631 \\
\bottomrule
\end{tabular}
\caption{\textbf{Transfer of demand across models} (AUROC, Eq.~\ref{eq:transfer}, pooled over failure types). Row: the model on which demand is measured. Column: the model whose untreated failures it predicts.}
\label{tab:transfer}
\end{minipage}
\end{table}

\subsection{Transfer within each failure type}

Within one failure type the relation between demand and failure has a single direction, so a one-covariate transfer fit loses nothing to cancellation. The nine source-to-target AUROCs per failure type range from $.56$ to $.68$ for negation, $.54$ to $.72$ for rare composition, and $.47$ to $.66$ for temporal. They are organized by \emph{target} and not by source: the columns with the strongest within-type signal (HunyuanVideo as target) are strongest from every source. No advantage for the same-model diagonal appears anywhere, and in several cells another model's demand predicts a target's failures as well as or better than the target's own (for example, in rare composition, HunyuanVideo$\to$Mochi $.627$ against Mochi$\to$Mochi $.606$).

\subsection{First-quarter demand, thresholds, and remaining checks}

\begin{table}[ht]
\begin{minipage}[t]{0.42\linewidth}
\centering
\small
\setlength{\tabcolsep}{4pt}
\begin{tabular}{lccc}
\toprule
Steps used & Mochi & Hunyuan & CogVX \\
\midrule
First quarter, $K{=}T/4$ & .876 & .871 & .857 \\
First half, $K{=}T/2$ & .967 & .940 & .972 \\
\bottomrule
\end{tabular}
\caption{\textbf{The early steps suffice.} Spearman $\rho$ between demand over the first $K$ steps and full-trajectory demand.}
\label{tab:prefix}
\end{minipage}\hfill
\begin{minipage}[t]{0.55\linewidth}
\centering
\small
\setlength{\tabcolsep}{4pt}
\begin{tabular}{lccc}
\toprule
Failure type & $D^{*}$ & 95\% CI & Slopes \\
\midrule
Rare composition & .589 & $[.486,\, 1.141]$ & $1.10 \to 0.14$ \\
Negation         & .212 & $[.168,\, .469]$  & $-1.50 \to -0.20$ \\
Temporal         & .316 & $[.273,\, .847]$  & $1.23 \to 0.19$ \\
\bottomrule
\end{tabular}
\caption{\textbf{Segmented fits of recovery on demand.} Threshold (breakpoint) $D^{*}$ with bootstrap 95\% CI, and the slopes before $\to$ after it. Per-method thresholds lie within the pooled bands.}
\label{tab:segmented}
\end{minipage}
\end{table}

On the embedding side, CLIP does not register what the judges register. In the three rare-composition cells where a method not designed for that failure type is applied, similarity to the common counterpart moves by $-.003$ under NEGATE, $-.000$ under TPD, and $+.004$ under scale-up, while judged recovery in those cells is $.476$, $.519$, and $.540$: scale-up slightly raises similarity to the counterpart even while compliance improves, because a global gain amplifies everything.
The difference measure of Eq.~\eqref{eq:realized} resolves where an untreated clip sits but not how a treated clip changed, which is why Fig.~\ref{fig:boundary} uses judged recovery.

For the check of the default prompt, clips generated from $\pd$ and scored by the unified judgment show the default in $8/8$ temporal, $6/8$ rare-composition, and $4/8$ negation cases ($n=8$ per failure type; the negation check is uninformative, see Sec.~\ref{subsec:default_operator}). In CLIP space, similarity to $\pd$ minus similarity to the source prompt is $+.028$ in rare composition, $+.005$ for temporal prompts, and $-.004$ in negation.

\end{document}